\documentclass[twocolumn,10pt]{article}
\usepackage[
top=1.78cm,
bottom=1.78cm,
left=1.4cm,
right=1.4cm,
headsep=0cm,
]{geometry}
\usepackage{etoolbox}
\usepackage[T1]{fontenc}
\usepackage{cite}
\usepackage{amsmath,amssymb,amsfonts}
\usepackage{algorithmic}
\usepackage{graphicx}
\usepackage{placeins}
\usepackage{setspace}
\usepackage{enumitem}
\usepackage{titlesec}
\usepackage{textcomp}
\usepackage[dvipsnames]{xcolor}
\usepackage[binary-units=true]{siunitx}
\usepackage[export]{adjustbox}
\usepackage{blindtext}
\usepackage{booktabs}
\usepackage{tikz}
\usepackage{multirow}
\usepackage{mwe,tikz}\usepackage[percent]{overpic}
\usepackage[font=footnotesize]{subcaption}
\usepackage[font=small, skip=5pt]{caption}
\usepackage[affil-it]{authblk}
\usetikzlibrary{shapes}
\usetikzlibrary{shapes.misc}
\tikzset{cross/.style={cross out, draw=black, minimum size=2*(#1-\pgflinewidth), inner sep=0pt, outer sep=0pt},
cross/.default={0.6mm}}

\DeclareSymbolFont{matha}{OML}{txmi}{m}{it}
\DeclareMathSymbol{\varv}{\mathord}{matha}{118}
\def\BibTeX{{\rm B\kern-.05em{\sc i\kern-.025em b}\kern-.08em
    T\kern-.1667em\lower.7ex\hbox{E}\kern-.125emX}}
    
\renewrobustcmd{\bfseries}{\fontseries{b}\selectfont}
\renewrobustcmd{\boldmath}{}
\newrobustcmd{\B}{\bfseries}

\AtBeginEnvironment{tabular}{\footnotesize}
\AtBeginEnvironment{abstract}{\itshape}
\AtBeginEnvironment{thebibliography}{\fontsize{8.5pt}{9.75pt}\selectfont}
\apptocmd{\thebibliography}{\setlength{\itemsep}{-4pt}}{}{}

\setlist{nosep}

\DeclareMathOperator*{\argmin}{arg\,min}

\makeatletter
\newcommand*\bigcdot{\mathpalette\bigcdot@{.5}}
\newcommand*\bigcdot@[2]{\mathbin{\vcenter{\hbox{\scalebox{#2}{$\m@th#1\bullet$}}}}}
\makeatother

\renewcommand{\thesection}{\Roman{section}}
\renewcommand{\thesubsection}{\Alph{subsection}} 
\renewcommand{\thesubsubsection}{\arabic{subsubsection}}
\titleformat{\section}{\normalfont\large\scshape\centering}{\thesection.}{0.5em}{}
\titleformat{\subsection}{\normalfont\itshape}{\thesubsection.}{0.5em}{}
\titleformat{\subsubsection}[runin]{\normalfont\itshape}{\thesubsubsection)}{0.5em}{}[:]
\titlespacing{\section}{0pt}{1.5ex plus 0.5ex minus .2ex}{1.0ex plus .2ex}
\titlespacing{\subsection}{0pt}{1.1ex plus 0.5ex minus .2ex}{0.5ex plus .2ex}
\titlespacing{\subsubsection}{\parindent}{0pt}{0.5em}
\makeatletter
\renewcommand{\p@section}{}
\renewcommand{\p@subsection}{\thesection-}
\renewcommand{\p@subsubsection}{\thesection-\thesubsection.}
\makeatother

\definecolor{light-gray}{gray}{0.95}
\AtBeginEnvironment{snugshade*}{\vspace{-\FrameSep}}
\AfterEndEnvironment{snugshade*}{\vspace{-\FrameSep}}

\makeatletter
\def\footnoterule{\relax%
  \kern-5pt
  \hbox to \columnwidth{\vrule width 0.45\columnwidth height 0.4pt\hfill}
  \kern4.6pt
  }
\makeatother

\makeatletter 
\newcommand\semiHuge{\@setfontsize\semiHuge{22.72}{29.5}}
\makeatother
\makeatletter
\patchcmd{\@maketitle}{\@title}{\vspace{-1cm}\semiHuge \@title}{}{}
\patchcmd{\@maketitle}{\@author}{\normalsize\@author}{}{}
\makeatother
\begin{document}
\title{Anomaly Detection in IR Images of PV Modules using Supervised Contrastive Learning}

\author[1]{Lukas Bommes}
\author[2,3]{Mathis Hoffmann}
\author[1]{Claudia Buerhop-Lutz}
\author[1]{Tobias Pickel}
\author[1]{Jens Hauch}
\author[1,3]{Christoph Brabec}
\author[2]{Andreas Maier}
\author[1]{Ian Marius Peters}
\affil[1]{Forschungszentrum Jülich GmbH, Helmholtz-Institute Erlangen-Nuremberg for Renewable Energies (HI ERN)}
\affil[2]{Pattern Recognition Lab, Department Informatik, Universität Erlangen-Nürnberg (FAU)}
\affil[3]{Institute Materials for Electronics and Energy Technology, Universität Erlangen-Nürnberg (FAU)\par\normalsize\normalfont{Correspondence to i.peters@fz-juelich.de}}

\date{}

\maketitle
\thispagestyle{empty} 

\begin{abstract}
\begin{spacing}{0.9}
Increasing deployment of photovoltaic (PV) plants requires methods for automatic detection of faulty PV modules in modalities, such as infrared (IR) images. Recently, deep learning has become popular for this. However, related works typically sample train and test data from the same distribution ignoring the presence of domain shift between data of different PV plants. Instead, we frame fault detection as more realistic unsupervised domain adaptation problem where we train on labelled data of one source PV plant and make predictions on another target plant. We train a ResNet\nobreakdash-$34$ convolutional neural network with a supervised contrastive loss, on top of which we employ a $k$\nobreakdash-nearest neighbor classifier to detect anomalies. Our method achieves a satisfactory area under the receiver operating characteristic (AUROC) of \SI{73.3}{\percent} to \SI{96.6}{\percent} on nine combinations of four source and target datasets with $2.92$ million IR images of which \SI{8.5}{\percent} are anomalous. It even outperforms a binary cross-entropy classifier in some cases. With a fixed decision threshold this results in \SI{79.4}{\percent} and \SI{77.1}{\percent} correctly classified normal and anomalous images, respectively. Most misclassified anomalies are of low severity, such as hot diodes and small hot spots. Our method is insensitive to hyperparameter settings, converges quickly and reliably detects unknown types of anomalies making it well suited for practice. Possible uses are in automatic PV plant inspection systems or to streamline manual labelling of IR datasets by filtering out normal images. Furthermore, our work serves the community with a more realistic view on PV module fault detection using unsupervised domain adaptation to develop more performant methods with favorable generalization capabilities.
\end{spacing}
\end{abstract}



\section{Introduction}

Solar photovoltaics (PV) has emerged as an important renewable energy source with a global installed capacity of \SI{627}{\giga\watt}$_\textrm{p}$ in $2020$ \cite{ren21.2020} that is projected to reach \SI{2840}{\giga\watt}$_\textrm{p}$ in $2030$ \cite{irena.2019}. PV modules are prone to defects due to aging, environmental influences or incorrect handling during installation. Defective modules pose safety hazards and reduce power output, yield, and profitability of a PV plant. Thus, regular inspection of PV plants is inevitable. As increasing plant sizes render manual inspection impractical, there is a recent surge in works on automatic inspection tools \cite{Bommes.2021, Aghaei.2015, Arenella.2017, Henry.2020b, Grimaccia.2017, Grimaccia.2018, Alsafasfeh.2018, Carletti.2019, Addabbo.2018, Deitsch.2016, Kim.2017, Dunderdale.2020, Oliveira.2019, Pierdicca.2018}, which use computer vision methods to automatically detect defective PV modules in modalities, such as aerial thermographic infrared (IR) images. 

The most recent methods frame fault detection as supervised classification and train a deep convolutional neural network with standard cross-entropy loss to classify different types of PV module faults in IR images \cite{Dunderdale.2020, Bommes.2021}. These methods achieve a high detection accuracy on the test dataset which is sampled from the same distribution as the training data. However, this setting ignores the fact that data distributions differ between plants, a problem known as domain shift. We find significant domain shift by examining $4.16$ million IR images from six different PV plants. Thus, we frame fault detection more realistically as unsupervised domain adaptation. Here, training is performed on labelled IR images of one source PV plant and predictions are made on another target PV plant for which no labels are available. This setting is more realistic as it takes domain shift into account. It is also more practical as training is performed only once, and no subsequent fine-tuning is needed when applying the fault detector to a new PV plant. Another challenge we address is the detection of unknown anomaly types which are present in the target dataset but not in the source dataset. This is generally known as open-set classification.

In this work, we develop a novel PV module anomaly detection method for IR images based on deep learning which addresses the aforementioned challenges. We train a ResNet\nobreakdash-$34$ convolutional neural network \cite{He.2016} with a supervised contrastive loss on labelled IR images of a source plant and use it to extract low-dimensional representations of the images. Based on these representations a $k$\nobreakdash-nearest neighbor ($k$\nobreakdash-NN) classifier detects anomalies in the target plant. By framing anomaly detection as supervised binary classification we follow a promising recent trend in the field \cite{Bergman.2020b, Ruff.2020b, Hendrycks.2019, Hendrycks.2019b}. Instead of performing active domain adaptation our method uses contrastive representations which are more informative and less domain-specific than representations learned by the standard cross-entropy loss \cite{Khosla.2020, Winkens.2020}. This also facilitates generalization beyond the training dataset and thus detection of unknown anomalies.

To summarize, our contributions are as follows:
\begin{itemize}
    \item We frame PV module fault detection as more realistic unsupervised domain adaptation problem where training is performed on one labelled source plant and anomalies are detected in another target PV plant.
    \item We introduce a domain-agnostic anomaly detection method based on contrastive representation learning and a binary $k$\nobreakdash-NN classifier which outperforms a binary cross-entropy classifier on some tasks and reliably detects unknown anomalies.
    \item We validate our method on nine combinations of four source and target datasets containing a total of $2.92$ million IR images.
\end{itemize}


\section{Related Works}
\label{sec:related_works}

In this section we briefly review related works on contrastive representation learning, domain adaptation, anomaly detection and PV module fault detection in IR images.

\subsection{Contrastive Representation Learning}
\label{sec:contrastive_learning}

Contrastive representation learning is a form of deep metric learning initially proposed by Hadsell et al.~\cite{Hadsell.2006}, which succeeds the older triplet \cite{Weinberger.2009} and $N$\nobreakdash-pair losses \cite{Sohn.2016}. For a good review see Le-Khac et al.~\cite{LeKhac.2020}. Contrastive representation learning uses deep neural networks to learn a low-dimensional feature space of high-dimensional data in which semantically similar samples are closer than semantically dissimilar ones. To this end, representations of a set of \emph{positive} samples are attracted and repulsed from the representations of all other (\emph{negative}) samples using for example the InfoNCE \cite{Oord.2018} or NT-Xent \cite{Chen.2020} loss. In the conventional self-supervised setting a single sample \cite{Wu.2018, Oord.2018}, and optionally perturbed versions of it \cite{Ye.2019, Chen.2020, He.2020, ChenXinlei.2020}, are used as positives. In the supervised setting all samples with the same class label (and optional perturbations) are positives \cite{Khosla.2020, ChenWei.2020, Kopuklu.2020}. Self-supervised contrastive representations discriminate individual samples. Supervised contrastive representations on the other hand discriminate classes by learning feature spaces in which samples are clustered based on their class membership. In our work we use contrastive representations because they are more informative than those learned with standard cross-entropy loss which retain only the minimum of information needed to discriminate training samples \cite{Khosla.2020, Winkens.2020}. This allows to extract discriminative features which are robust against domain shift and generalize to unseen classes.

\subsection{Domain Adaptation}
\label{sec:domain_adaption}

Domain adaptation addresses the problem of learning transferable representations without the need for large amounts of labelled training data. For a good overview we refer the reader to the surveys by Wang et al.~\cite{Wang.2018} and Zhao et al.~\cite{Zhao.2020}. Our problem corresponds to unsupervised domain adaptation where we learn representations on labelled data of a source domain that generalize to an unlabelled target domain. Many domain adaptation methods estimate and minimize the discrepancy between source and target domain by means of loss functions, such as Maximum Mean Discrepancy \cite{Long.2015, Rozantsev.2019, Zhu.2019}, L$2$- or cosine distance \cite{Guo.2020, Rakshit.2019}, Rényi divergence \cite{Hoffman.2018} or KL\nobreakdash-divergence \cite{Zhuang.2015}. Recently, contrastive losses have been used as well \cite{Kang.2019, Dai.2020, Park.2020}. Aligning source and target representations this way improves performance when classifying images \cite{Zhu.2021} or detecting anomalies \cite{Yang.2019} in the target domain. While our method does not use any domain adaptation loss, it solves the same problem by using more informative and thus less domain-specific contrastive representations.

\subsection{Anomaly Detection}
\label{sec:anomaly_detection}

Anomaly detection (AD) aims at identifying anomalous data samples which deviate from the majority of normal samples. This relates to our dataset which contains mostly normal PV modules and only a small fraction of faulty modules. For a good overview of recent deep learning-based AD methods we refer to the surveys by Pang et al.~\cite{Pang.2021}, Bulusu et al.~\cite{Bulusu.2020} and Chalapathy et al.~\cite{Chalapathy.2019}. Most deep AD methods learn representations of normal data using autoencoders \cite{ChenJinghui.2017, An.2015}, generative adversarial networks \cite{Akcay.2019, Zenati.2018}, one-class losses \cite{Ruff.2020, Ruff.2018}, self-supervised learning \cite{Golan.2018, Wang.2019, Hendrycks.2019, Bergman.2020} or metric learning \cite{Perera.2019, Yilmaz.2020} and identify anomalies by a high reconstruction error or a large distance to the normal representations. Recently, (self\nobreakdash-)supervised contrastive learning has gained popularity for learning representations for AD \cite{Sohn.2021, Winkens.2020, Tack.2020, Kopuklu.2020}. Some works also explored the use of domain adaptation for anomaly detection \cite{ChenJixu.2014, Yang.2019, Yamaguchi.2019, Kumagai.2019}.

Many AD methods assume an unlabelled training dataset containing mostly normal samples and a few anomalies. If labelled anomalies are available AD can also be formulated as (semi\nobreakdash-)supervised binary classification and achieve state-of-the-art performance \cite{Ruff.2020b, Hendrycks.2019, Hendrycks.2019b}. Similary, using a supervised $k$\nobreakdash-NN classifier on embeddings of a ResNet, which is pretrained on ImageNet with cross-entropy loss, outperforms many other AD methods \cite{Bergman.2020b}.

Building on this, our work formulates AD as supervised binary classification with a $k$\nobreakdash-NN classifier. As opposed to the other works we use contrastive representations and perform anomaly detection in a target domain which differs from the source domain and does not contain any labelled examples.

\subsection{PV Module Fault Detection}

Until recently, PV module faults were detected as hot regions in IR images using classical computer vision algorithms, such as segmentation by intensity thresholding \cite{Aghaei.2015, Arenella.2017, Henry.2020b, Grimaccia.2017, Grimaccia.2018}, iterative growth of segmentation masks \cite{Alsafasfeh.2018, Carletti.2019} or template matching \cite{Addabbo.2018}. Downside of these methods is their dependence on heuristics and manual priors, the need for extensive manual tuning and poor generalization to unseen imagery. The extraction of hand-crafted image features and detection of outliers by statistical tests \cite{Deitsch.2016, Kim.2017} or classification with a SVM or Random Forest \cite{Dunderdale.2020} is slightly more robust. Recently, deep learning has shown promising results in overcoming the problems of classical algorithms \cite{Bommes.2021, Dunderdale.2020, Oliveira.2019, Pierdicca.2018, Mayr.2020}. Typically, fault detection is performed as a supervised classification in which deep convolutional networks, such as ResNet, MobileNet \cite{Howard.2017} or VGG \cite{Simonyan.2014}, are trained with standard cross-entropy loss to distinguish a predefined set of fault classes. To the best of our knowledge, related works in the field have neither addressed the problem of domain shift nor the detection of unknown anomaly classes.


\section{Dataset}
\label{sec:dataset}

We use an extended version of the dataset from our previous work \cite{Bommes.2021}. It consists of $4.16$ million IR images showing $105546$ PV modules from six different PV plants, which were acquired under clearsky conditions and solar irradiance above \SI{700}{\watt\per\square\meter}. Note, that we name the PV plants A to G in accordance to our previous work. We omit plant D as it contains thin-film modules instead of crystalline silicon modules like the other plants.

\begin{figure}[tbp]
     \captionsetup[subfigure]{font=scriptsize, aboveskip=2pt, belowskip=0pt, labelformat=empty, justification=centering}
     \newcommand\sizefactor{0.153}
     \centering
     \begin{subfigure}[t]{\sizefactor\columnwidth}
         \centering
         \includegraphics[width=\textwidth]{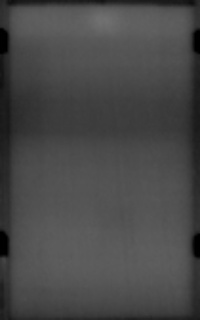}
        \caption{Normal}
     \end{subfigure}
     \begin{subfigure}[t]{\sizefactor\columnwidth}
         \centering
         \includegraphics[width=\textwidth]{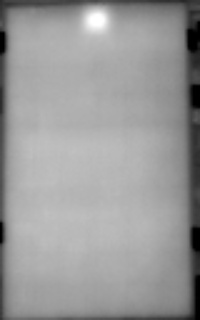}
         \caption{Mh: Module open-circuit}
     \end{subfigure}
     \begin{subfigure}[t]{\sizefactor\columnwidth}
         \centering
         \includegraphics[width=\textwidth]{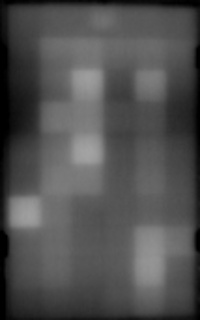}
         \caption{Mp: Module short-circuit}
     \end{subfigure}
     \begin{subfigure}[t]{\sizefactor\columnwidth}
         \centering
         \includegraphics[width=\textwidth]{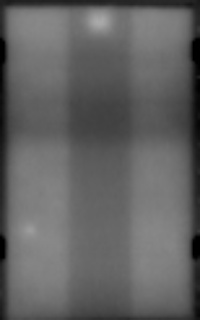}
         \caption{Sh: Substring open-circuit}
     \end{subfigure}
     \begin{subfigure}[t]{\sizefactor\columnwidth}
         \centering
         \includegraphics[width=\textwidth]{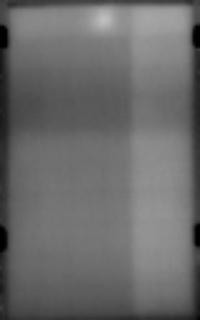}
         \caption{Sh}
     \end{subfigure}
     \begin{subfigure}[t]{\sizefactor\columnwidth}
         \centering
         \includegraphics[width=\textwidth]{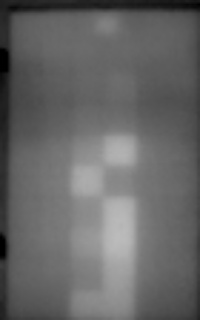}
         \caption{Sp: Substring short-circuit}
     \end{subfigure}
     \par\smallskip
     \begin{subfigure}[t]{\sizefactor\columnwidth}
         \centering
         \includegraphics[width=\textwidth]{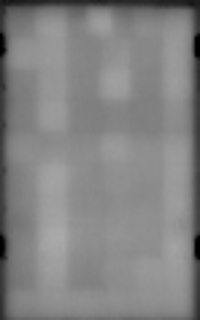}
         \caption{Pid: Module PID}
     \end{subfigure}
     \begin{subfigure}[t]{\sizefactor\columnwidth}
         \centering
         \includegraphics[width=\textwidth]{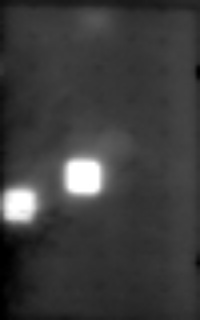}
         \caption{Cm+: Multi. hot cells}
     \end{subfigure}
     \begin{subfigure}[t]{\sizefactor\columnwidth}
         \centering
         \includegraphics[width=\textwidth]{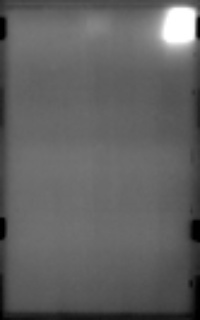}
         \caption{Cs+: Single hot cell}
     \end{subfigure}
     \begin{subfigure}[t]{\sizefactor\columnwidth}
         \centering
         \includegraphics[width=\textwidth]{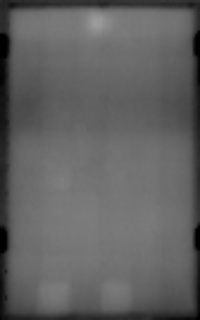}
         \caption{C: Warm cell(s)}
     \end{subfigure}
     \begin{subfigure}[t]{\sizefactor\columnwidth}
         \centering
         \includegraphics[width=\textwidth]{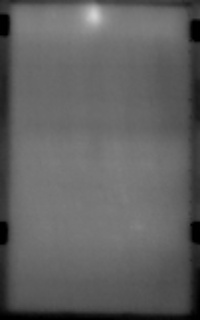}
         \caption{D: Diode overheated}
     \end{subfigure}
     \begin{subfigure}[t]{\sizefactor\columnwidth}
         \centering
         \includegraphics[width=\textwidth]{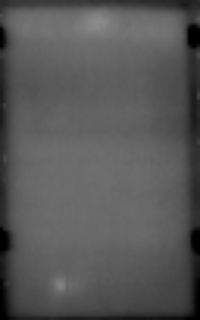}
         \caption{Chs: Hot spot}
     \end{subfigure}
        \caption{Exemplary IR images of a normal and ten different types of anomalous PV modules in our dataset. Temperature ranges from \SI{30}{\celsius} (black) to \SI{60}{\celsius} (white). All images except for class Cm+ show plant A. The figure is taken from our previous work \cite{Bommes.2021}.}
        \label{fig:dataset_fault_patches}
\end{figure}

Images are cropped from IR videos of a drone-mounted DJI Zenmuse XT$2$ camera and rectified to remove perspective distortion. Due to redundancies in the video, there are on average $39.4$ images of each PV module which serve as multiple augmented views. Each image is labelled by an expert either as containing a normal module or a module with one out of the ten typical faults shown in fig.~\ref{fig:dataset_fault_patches}. While our method makes only a binary distinction between normal and anomalous modules, fine-grained fault labels are used to evaluate our method.

Tab.~\ref{tab:dataset_composition} and tab.~\ref{tab:dataset_composition_fine_grained} show the distribution of anomaly classes in our dataset. To ensure a realistic setting, we do not balance the numbers of normal and anomalous images. For our experiments we use only data of plants A, B, E and F as plant C contains very few anomalies and ground truth labels of plant G were not obtained by an expert. Each dataset is split each into \SI{70}{\percent} train and \SI{30}{\percent} test data. Here, we ensure that images of the same PV module do not occur in both train and test set.
 
\begin{figure}[tbp]
     \captionsetup[subfigure]{aboveskip=2pt, belowskip=0pt, justification=centering}
     \centering
     \begin{subfigure}[t]{0.48\columnwidth}
         \centering
         \begin{overpic}[abs, unit=1mm, width=\linewidth]{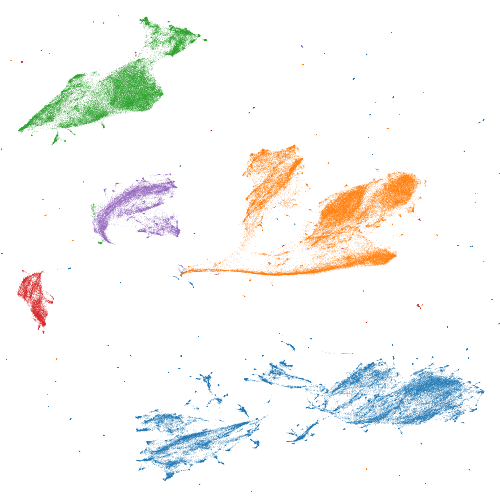}
         \put(30, 2){A}
         \put(30, 30){B}
         \put(5, 12){E}
         \put(5, 25){F}
         \put(17, 36){G}
         \end{overpic}
         \caption{}\label{fig:dataset_umap_embedding_plants}
     \end{subfigure}
     \begin{subfigure}[t]{0.48\columnwidth}
         \centering
         \includegraphics[width=\linewidth]{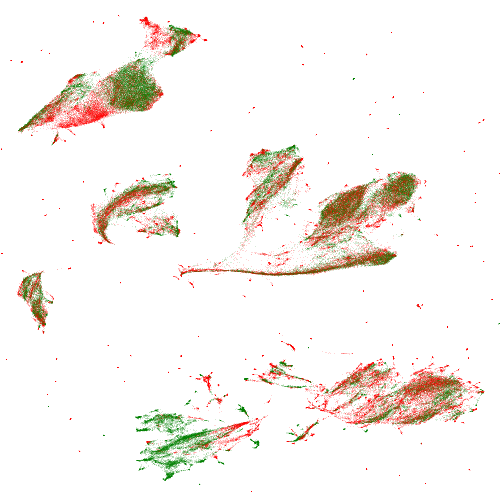}
         \caption{}\label{fig:dataset_umap_embedding_classes}
     \end{subfigure}
        \caption{Projection of our dataset obtained by UMAP (with $50$ neighbors per sample and minimum distance of $0.1$). Colors in (a) indicate the PV plant, which reveals the domain shift between different plants. In (b) normal and anomalous samples are colored green and red, respectively. UMAP is applied directly to the flattened images, which are preprocessed as in sec. \ref{sec:image_preprocessing}. For better visualization, normal samples are subsampled to match the number of anomalous samples.}
        \label{fig:dataset_umap_embedding}
\end{figure}
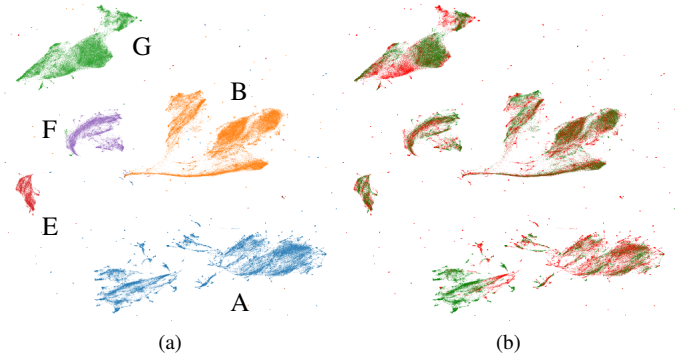

\begin{figure}[tbp]
     \captionsetup[subfigure]{aboveskip=2pt, belowskip=0pt, labelformat=empty, justification=centering}
     \newcommand\commonheight{23.04mm} 
     \centering
     \begin{subfigure}[t]{17.568mm}
         \centering
         \includegraphics[height=\commonheight, width=\linewidth]{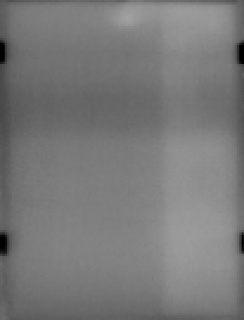}
        \caption{Plant A}
     \end{subfigure}
     \begin{subfigure}[t]{14.4mm}
         \centering
         \includegraphics[height=\commonheight, width=\linewidth]{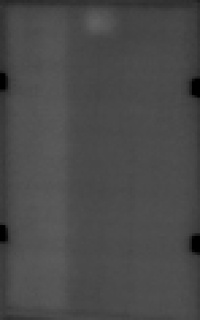}
         \caption{Plant B}
     \end{subfigure}
     \begin{subfigure}[t]{14.184mm}
         \centering
         \includegraphics[height=\commonheight, width=\linewidth]{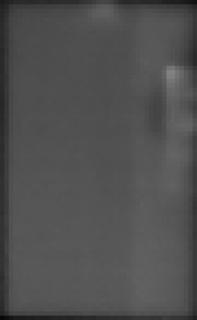}
         \caption{Plant C}
     \end{subfigure}
     \begin{subfigure}[t]{11.736mm}
         \centering
         \includegraphics[height=\commonheight, width=\linewidth]{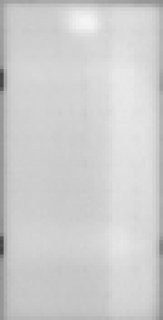}
         \caption{Plant E}
     \end{subfigure}
     \begin{subfigure}[t]{13.248mm}
         \centering
         \includegraphics[height=\commonheight, width=\linewidth]{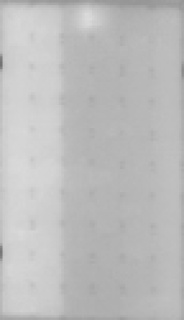}
         \caption{Plant F}
     \end{subfigure}
     \begin{subfigure}[t]{14.184mm}
         \centering
         \includegraphics[height=\commonheight, width=\linewidth]{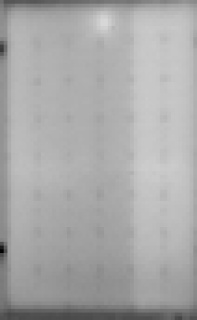}
         \caption{Plant G}
     \end{subfigure}
        \caption{IR images differ between the PV plants in our dataset due to different ambient conditions, camera positions, as well as module and cell type. Shown are modules with Sh anomaly. The original aspect ratio is preserved and temperature ranges from \SI{15}{\celsius} (black) to \SI{50}{\celsius} (white).}
        \label{fig:dataset_plant_patches}
\end{figure}

Fig. \ref{fig:dataset_umap_embedding} shows UMAP embeddings \cite{McInnes.2018} of our dataset. Here, images form distinctive clusters or \emph{domains} depending on the PV plant they originate from. This domain shift has various reasons, such as differences in ambient conditions, camera position, as well as module and cell type. For most plants, we additionally observe sub-domains which correspond to different rows of vertically stacked modules. Fig.~\ref{fig:dataset_plant_patches} shows an exemplary patch for each plant clearly revealing differences. We also found that different module orientations in the images lead to domain shift. To account for this we rotate all images so that module junction boxes are always at the top edge.

\begin{table}[tbp]
\centering
\caption{Numbers of normal and anomalous IR images in our dataset.}
\label{tab:dataset_composition}
\begin{tabular}{l
S[table-format=6.0]
S[table-format=6.0]
S[table-format=6.0]
S[table-format=6.0]
S[table-format=6.0]
S[table-format=7.0]}
\toprule
{Class} & \multicolumn{6}{c}{Plant}\\\cmidrule(lr){2-7}
        & {A} & {B} & {C} & {E} & {F} & {G}\\\midrule
Normal & 864394 & 869957 & 135342 & 751261 & 185613 & 1043216\\
Anomalous & 107786 & 98206 & 306 & 15174 & 25841 & 63383 \\\midrule
Normal (\si{\percent}) & {88.91} & {89.86} & {99.77} & {98.02} & {87.8} & {94.27} \\
Anomalous (\si{\percent}) & {11.09} & {10.14} & {0.23} & {1.98} & {12.2} & {5.73} \\
\bottomrule
\end{tabular}
\end{table}

\begin{table}[tbp]
\centering
\caption{Numbers of anomalous IR images per underlying fault class.}
\label{tab:dataset_composition_fine_grained}
\begin{tabular}{l
S[table-format=5.0]
S[table-format=5.0]
S[table-format=3.0]
S[table-format=5.0]
S[table-format=5.0]
S[table-format=5.0]}
\toprule
{Class} & \multicolumn{6}{c}{Plant}\\\cmidrule(lr){2-7}
        & {A} & {B} & {C} & {E} & {F} & {G}\\\midrule
Mh  &   212 & 33129 & 112 &     0 &    38 & 19968\\
Mp  &    74 &   185 & 151 &   272 &    62 &    26\\
Sh  &  2421 &  2594 &  43 &    73 &    13 &   145\\
Sp  &   360 &   328 &   0 &  1802 &   217 &  1573\\
Pid & 40422 & 23174 &   0 &     0 &     0 &     0\\
Cm+ &    26 &   388 &   0 &   477 &   352 &     0\\
Cs+ &   468 &  1651 &   0 &   582 &  1348 &     0\\
C   & 36955 & 28174 &   0 & 11618 & 23539 &   256\\
D   & 24891 &    66 &   0 &     0 &   197 & 41210\\
Chs &  1957 &  8517 &   0 &   350 &    75 &   205\\\bottomrule
\end{tabular}
\end{table}

\section{method}
\label{sec:method}

\begin{figure*}[tbp]
\centering
\definecolor{custom-orange}{rgb}{1.0, 0.42745098, 0.0}
\definecolor{custom-green}{rgb}{0.0, 0.501960784, 0.0}
\begin{overpic}[abs, unit=1mm, scale=1.0, tics=5]{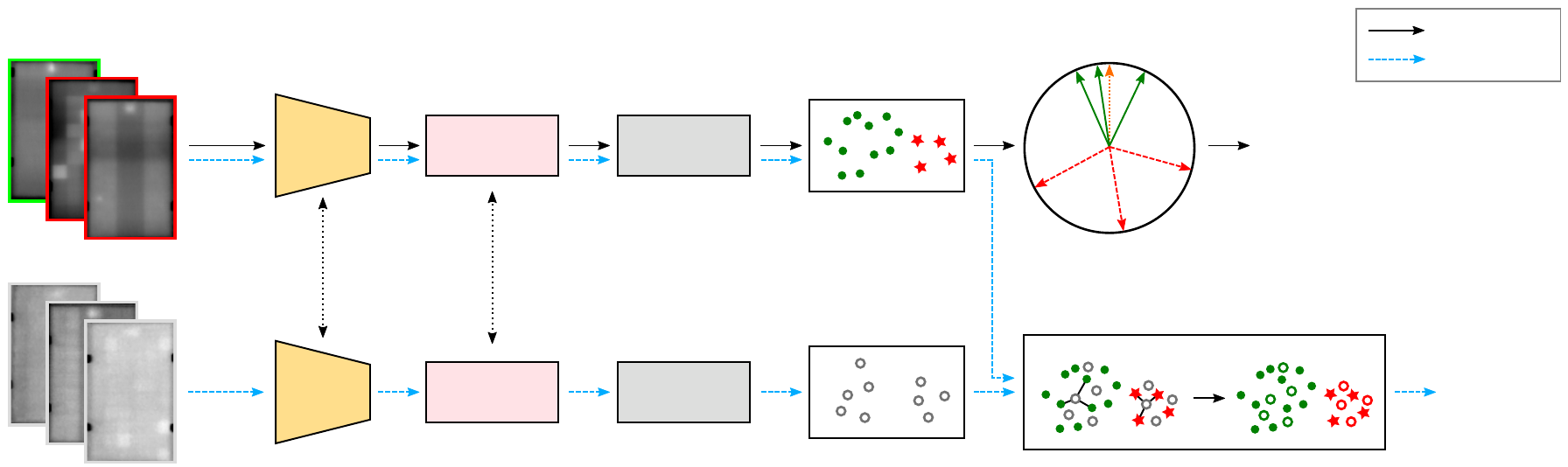}
    \put (0.0, 54.9) {\parbox[t][8mm][c]{20.0mm}{\footnotesize \centering Source Images\\$+$ Labels}}
    \put (0.0, 27.0) {\parbox[t][8mm][c]{20.0mm}{\footnotesize \centering Target Images}}
    \put(22.234, 39.5){\footnotesize $x_i^S$, $y_i^S$}
    \put(24.5, 10.5){\footnotesize $x_i^T$}
    \put (31.925, 41.0) {\parbox[t][8mm][c]{11.153mm}{\footnotesize \centering Encoder\\$f_\theta{\left( \cdot \right)}$}}
    \put (31.925, 12.359) {\parbox[t][8mm][c]{11.153mm}{\footnotesize \centering Encoder\\$f_\theta{\left( \cdot \right)}$}}
    \put (49.389, 40.897) {\parbox[t][8mm][c]{15.553mm}{\footnotesize \centering Projection\\Head $h_\psi{\left( \cdot \right)}$}}
    \put (49.389, 12.256) {\parbox[t][8mm][c]{15.553mm}{\footnotesize \centering Projection\\Head $h_\psi{\left( \cdot \right)}$}}
    \put (71.623, 41.197) {\parbox[t][8mm][c]{15.553mm}{\footnotesize \centering \linespread{0.85} \selectfont L$2$-\\normalize}}
    \put (71.623, 12.556) {\parbox[t][8mm][c]{15.553mm}{\footnotesize \centering \linespread{0.85} \selectfont L$2$-\\normalize}}
    \put (92.413, 50.5) {\parbox[t][8mm][c]{21.214mm}{\footnotesize \centering Source\\Embeddings $z_i^S$}}
    \put (92.413, 21.859) {\parbox[t][8mm][c]{21.214mm}{\footnotesize \centering Target\\Embeddings $z_i^T$}}
    \put (129.646, 21.5) {\parbox[t][8mm][c]{20.399mm}{\footnotesize \centering $k$-NN Classifier}}
    \put (121.385, 54.9) {\parbox[t][8mm][c]{14.877mm}{\footnotesize \centering Unit\\Hypersphere}}
    \put (146.8, 41.0) {\parbox[t][8mm][c]{16.091mm}{\footnotesize Supervised\\Contrastive\\Loss $\mathcal{L}_{\textrm{AD}}$}}
    \put (168.0, 12.0) {\parbox[t][8mm][c]{13.107mm}{\footnotesize Target\\Labels $y_i^T$}}
    \put (166.750, 54.5) {\parbox[t][8mm][c]{13.503mm}{\footnotesize Training}}
    \put (166.750, 51.2) {\parbox[t][8mm][c]{13.503mm}{\footnotesize Prediction}}
    \put (119.928, 11.7) {\footnotesize $1$.}
    \put (142.594, 11.7) {\footnotesize $2$.}
    \put (41.999, 27.5) {\parbox[t][8mm][c]{11.071mm}{\footnotesize \centering Shared\\Weights}}
    \put (129.0, 42.0) {\footnotesize \color{custom-orange} $\bar{z}$}
    \put (129.0, 42.0) {\footnotesize \color{custom-orange} $\bar{z}$}
    \put (95.0, 39.5) {\footnotesize \color{custom-green} $\mathcal{N}$}
    \put (107.0, 39.5) {\footnotesize \color{red} $\mathcal{A}$}
    \put (132.0, 38.0) {\footnotesize $z_i^S$}
\end{overpic}
\caption{Overview of our method for detection of anomalous PV modules in IR images of a \emph{target} PV plant based on labelled samples of a \emph{source} PV plant. Low dimensional embeddings of both source and target images are extracted by means of contrastive representation learning. A $k$\nobreakdash-NN classifier predicts target labels based on the labels of neighboring source images in the embedding space.}
\label{fig:overview_image} 
\end{figure*}
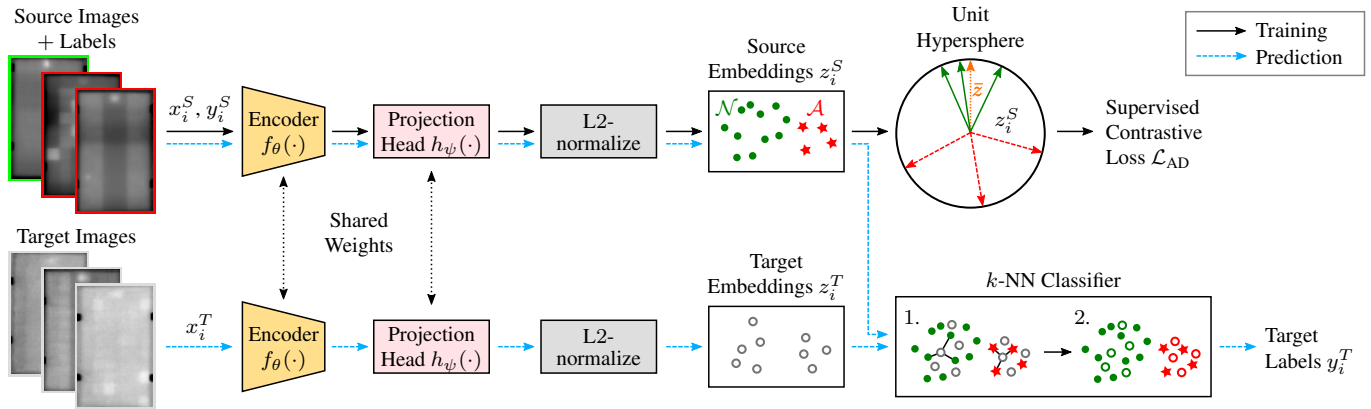

The aim of our method is to predict binary labels $\left\lbrace \hat{y}_i^T \right\rbrace_{i=1\ldots N^T}$ for $N^T$ IR images $\left\lbrace x_i^T \right\rbrace_{i=1\ldots N^T}$ of a \emph{target} PV plant, depending on whether a normal or an anomalous PV module is shown. While we have no labelled examples for this PV plant, we have a set of $N^S$ binary labelled images $\left\lbrace \left( x_i^S, y_i^S \right) \right\rbrace_{i=1\ldots N^S}$ of at least one other \emph{source} PV plant. Typically, there is a domain shift between source and target images and the distribution of anomaly classes between source and target can differ significantly. The target data can even contain unknown anomalies, which are not present in the source data. Our method shown in fig.~\ref{fig:overview_image} overcomes these challenges by i) learning informative and domain-agnostic representations with a supervised contrastive loss and ii) detecting unknown anomalies on top of the representations with a $k$\nobreakdash-NN classifier.

\subsection{Supervised Contrastive Representation Learning}

As indicated by fig.~\ref{fig:dataset_umap_embedding_classes} we observe that IR images form clusters depending on the PV plant they originate from. However, they do not form clusters of normal and anomalous images. We employ representation learning to compute a low-dimensional embedding of the IR images which forms distinctive clusters of normal and anomalous images and reduces clustering by plants. Extraction of low-dimensional embeddings from the high-dimensional IR images is also needed to make anomaly detection computationally tractable. Instead of using hand-crafted features, we employ deep neural networks and a supervised contrastive loss to learn a suitable embedding end-to-end. Specifically, we use a convolutional encoder $f_\theta{\left( \cdot \right)}$ and a fully connected projection head $h_\psi{\left( \cdot \right)}$ to extract a $d$\nobreakdash-dimensional embedding vector $v_i^S \in \mathbb{R}^d$ from each source image $x_i^S$
\begin{equation}
\label{eq:feature_extraction}
    v_i^S = h_\psi{\left( f_\theta{\left( x_i^S \right) }\right)} \textrm{.}
\end{equation}
Several related works use a projection head to improve representational power of the encoder embeddings \cite{Chen.2020, Grill.2020, ChenXinlei.2020}. We follow this architecture choice. Note, however that the effect on the encoder embeddings is less relevant in our case as we use the embeddings after the projection head instead for anomaly detection.

After encoding, each embedding vector is normalized to unit L$2$\nobreakdash-norm
\begin{equation}
\label{eq:l2_normalization}
    z_i^S = v_i^S / \left\lVert v_i^S \right\rVert_2 \textrm{.}
\end{equation}
Iterative stochastic gradient descent is performed on embeddings of randomly shuffled batches of $N$ labelled source images $\left\lbrace x_i^S, y_i^S \right\rbrace_{i=1\ldots N}$ to compute suitable network parameters
\begin{equation}
\label{eq:representation_learning_training_framework}
    \left\lbrace \theta^*, \psi^* \right\rbrace = \argmin_{\theta, \psi} \mathcal{L}_{\textrm{AD}}{\left( z_i^S, y_i^S \right)}
\end{equation}
where $\mathcal{L}_{\textrm{AD}}$ is a supervised contrastive loss with the following form of a non-parametric softmax classifier \cite{LeKhac.2020}
\begin{equation}
        \mathcal{L}_{\textrm{AD}}{\left( z_i^S, y_i^S \right)} = - \frac{1}{\left| \mathcal{N} \right|} \sum_{i \in \mathcal{N}} \log{\frac{\exp{ \left( z_i^S \bigcdot \bar{z}^S / \tau \right)}}{\sum_{j \in \mathcal{N} \cup \mathcal{A}} \exp{ \left( z_j^S \bigcdot \bar{z}^S / \tau \right)}}} \textrm{.}
\end{equation}
Here, the $\bigcdot$ symbol denotes the dot product of two vectors and $\tau \in \mathbb{R}^+$ is a scalar temperature hyperparameter as used by Wu et al.~\cite{Wu.2018} and He et al.~\cite{He.2020}. We set $\tau = 0.1$ for all experiments. Further, $\mathcal{N}$ and $\mathcal{A}$ denote the indices of all normal and anomalous embeddings in the current batch and $\bar{z}^S \in \mathbb{R}^d$ is the mean vector of all normal embeddings
\begin{equation}
    \bar{z}^S = \frac{1}{\left| \mathcal{N} \right|} \sum_{i \in \mathcal{N}} z_i^S \textrm{.}
\end{equation}

This loss is based on the normalized temperature scaled cross-entropy loss \cite{Chen.2020, Khosla.2020} and the central contrastive loss \cite{ChenWei.2020}. Intuitively, it pulls all normal samples in the batch towards the normal mean vector and pushes the anomalies away. While this causes formation of a single cluster of normal IR images in embedding space, anomalies can potentially form multiple clusters depending on the underlying anomaly class. Note, that pulling each normal sample towards the normal mean embedding has the same effect as pulling all pairs of normal embeddings towards each other. We use the first variant as it is easier to implement. 

\subsection{Anomaly Detection with a $k$\nobreakdash-NN Classifier}

The anomaly detection stage predicts for each target image $x_j^T$ whether it shows a normal or an anomalous PV module using a $k$\nobreakdash-NN classifier on top of the learned representations. First, the trained base encoder and projection head are used to compute the embeddings $\left\lbrace z_i^S \right\rbrace_{i=1\ldots N^S}$ of all source images as in eq.~\ref{eq:feature_extraction} and eq.~\ref{eq:l2_normalization}. This needs to be done only once, as the embeddings are persisted in memory. Similarly, the target embedding $z_j^T$ is computed. Now, the $k$ source embeddings nearest to the target embedding in terms of Euclidean distance are obtained. We denote them as $\mathcal{N}_k$. As all embeddings have unit L$2$\nobreakdash-norm using Euclidean distance is equivalent to using cosine distance. The final prediction $\hat{y}_j^T$ for the target image is made by aggregating the labels of the images in $\mathcal{N}_k$. If the fraction of anomalies in $\mathcal{N}_k$ exceeds the specified threshold $\delta$, the target image is predicted to contain an anomalous PV module. Later, in sec. \ref{sec:hyperparameter_selection_for_the_knn_classifier} we will determine optimal settings for the hyperparameters $k$ and $\delta$.

We also tried using temperature-scaled cosine distance $\exp{\left(z_j^T \bigcdot z_i^S / \tau \right)}$ and distance-weighted label aggregation for prediction as in Wu et al.~\cite{Wu.2018}. However, we did not observe a large impact on the predictions.

While in theory it is computationally expensive to compare each target embedding with all source embeddings, we do not observe this to be a bottleneck in practice for our dataset sizes. A possible workaround for significantly larger datasets is to perform $k$\nobreakdash-means clustering on the source embeddings, and to use only the cluster centroids for distance computations \cite{Goldberger.2004, Bergman.2020b}.

\subsection{Implementation Details}
\label{sec:implementation_details}

\subsubsection{Network Architecture}

We employ a randomly initialized ResNet\nobreakdash-$34$ without the final classification layer as convolutional encoder $f_\theta{\left( \cdot \right)}$. We add a $2$D global average pooling layer \cite{Huang.2017, He.2016} as final layer which outputs a $512$\nobreakdash-dimensional vector for each input image in the batch. The projection head $h_\psi{\left( \cdot \right)}$ is implemented by two randomly initialized fully-connected layers with $512$ and $128$ outputs, respectively, where the first layer is followed by a ReLU activation. Thus, the dimensionality of embeddings after the projection head is $d = 128$.

\subsubsection{Image Preprocessing}
\label{sec:image_preprocessing}

Prior to feature extraction each $16$\nobreakdash-bit grayscale IR image is converted to Celsius scale, normalized to the interval $\left[ 0, 255 \right]$ using the minimum and maximum temperature value in the image, converted to $8$\nobreakdash-bit and resized to $64 \times 64$ pixels. Each image is standardized by subtracting the dataset mean and dividing by the dataset standard deviation. To account for the domain shift, we compute a separate mean and standard deviation for each PV plant. As ResNet expects an RGB image as input, we finally stack three copies of the grayscale image along the channel-direction.

\subsubsection{Training}

We train all models for $110000$ steps using stochastic gradient descent with momentum $0.9$ and weight decay $5\times10^{-4}$ \cite{Sutskever.2013, Loshchilov.2018}. The initial learning rate $\eta_0 = 6\times10^{-2}$ is decayed in each step following the Cosine Annealing strategy $\eta = \eta_0/2 \left(1 + \cos{\left( p \pi \right)} \right)$ where $p \in \left[ 0, 1 \right]$ is the training progress \cite{Loshchilov.2017}. We train with $16$\nobreakdash-bit precision and batch size $128$ which is the maximum trainable on our hardware. We believe larger batch sizes can benefit contrastive representation learning as reported in similar works \cite{Khosla.2020, Chen.2020, He.2020}. During training, we augment both source and target images independently from another by random up\nobreakdash-down and left\nobreakdash-right flips and random rotation by multiples of \SI{90}{\degree}. All images in a batch are augmented identically.

\subsubsection{Hardware and Software}

All models are trained on a desktop workstation with an Intel i$9$\nobreakdash-$9900$K, \SI{64}{\giga\byte} RAM and a GeForce RTX $2080$ Ti running Ubuntu $20.04$ LTS, Python $3.6.9$, PyTorch $1.7.1$ and PyTorch Lightning $1.1.5$.


\section{Experiments \& Results}

In the following, we perform a quantitative analysis of our method and compare it against a binary cross-entropy classifier.

\subsection{Evaluation Protocol}

As common in anomaly detection, we evaluate all our models in terms of the area under the receiver operating characteristic (AUROC) and the average precision score (AP) \cite{Golan.2018, Bergman.2020b, Hendrycks.2019b}. AUROC is obtained by plotting the true positive rate $\textrm{TPR} = \textrm{TP} / (\textrm{TP} + \textrm{FN})$ over the false positive rate $\textrm{FPR} = \textrm{FP} / (\textrm{FP} + \textrm{TN})$ at various decision thresholds $\delta$ and integrating the resulting curve. Here, $\textrm{TP}$ and $\textrm{TN}$ denote the numbers of correctly classified anomalous and normal images, $\textrm{FP}$ is the number of normal images misclassified as anomalous and $\textrm{FN}$ the number of anomalous images classified as normal. 

Similarly, the AP is obtained from the precision-recall curve which plots precision $\textrm{P} = \textrm{TP} / (\textrm{TP} + \textrm{FP})$ over recall $\textrm{R} = \textrm{TP} / (\textrm{TP} + \textrm{FN})$) at different decision thresholds. The AP summarizes the curve as the weighted mean of precisions achieved at each threshold $\textrm{AP} = \sum_{i=1}^n \left( \textrm{R}_i - \textrm{R}_{i-1} \right) \textrm{P}_i$.

While AUROC takes both the normal class and the anomalous class into account, AP puts more emphasis on the anomalies \cite{Saito.2015}. Both AUROC and AP do not depend on a specific decision threshold $\delta$. Instead, they measure classification performance over the entire spectrum of threshold values. This makes them more informative than other metrics, such as classification accuracy or F$1$\nobreakdash-score, which are computed at a single threshold value. Because of this, AUROC and AP enable a fair comparison of different methods, which can depend differently on the decision threshold.

In the following, each model is trained on a source dataset S (train split) and evaluated on a target dataset T (train split), which we refer to as \emph{task} S \textrightarrow{} T. As mentioned in sec.~\ref{sec:dataset}, only the data of PV plants A, B, E and F is used. When we train and evaluate on the same PV plant, we use the source test split for evaluation and refer to it as A', B', E' or F'. We train each model three times with different random seeds and report the mean of AUROC and AP.

\subsection{Model Selection}

In the following experiments, we compute AUROC and AP after each training epoch and report the best values obtained. In practice this is not feasible as target labels are unknown. Thus, we use labelled data of a second PV plant as validation dataset and report the target AUROC (AP) for the epoch at which the highest validation AUROC (AP) is achieved.

Sun et al.~\cite{Sun.2020} proposed to use the cosine distance between the mean source and target embeddings for model selection. However, in our experiments this did not correlate well to the target metrics.

\subsection{Results of the Contrastive $k$\nobreakdash-NN Classifier}
\label{sec:results_of_the_contrastive_knn_classifier}

\begin{figure}[tbp]
\centering
\includegraphics[scale=0.5]{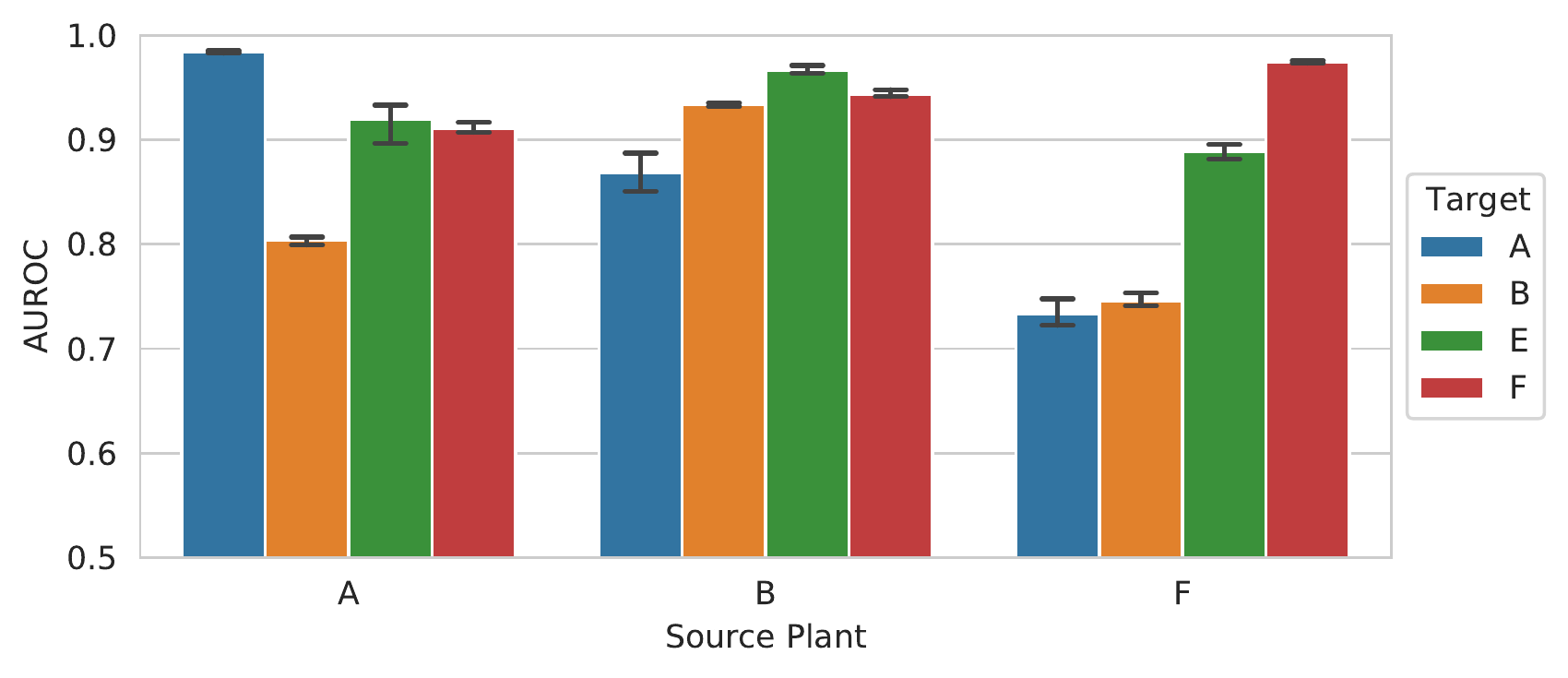}
\caption{Best target AUROC of our method when trained on different source PV plants. When source and target plant are identical we evaluate on the target test split otherwise on the target train split. Error bars indicate the \SI{95}{\percent} confidence interval over three runs.}
\label{fig:source_vs_target_plants} 
\end{figure}

We train and evaluate our method on various tasks and report the best target AUROC scores in fig.~\ref{fig:source_vs_target_plants}. All scores are above \SI{70}{\percent} and thus well above the \SI{50}{\percent} of a random guess. When training and evaluating on the same plant AUROC scores are generally higher, as there is no domain shift between train and test data. The results suggest that the choice of source plant has a considerable effect on the achievable target AUROC. For example, plant B is a better source plant than A and plant A is better than F. Plant F is most likely the worst source plant because its dataset is $4.6$ times smaller than that of plants A and B. However, plant A and B are similar in sample count and distribution of anomaly classes. Hence, it is interesting that plant B is a better source plant. This indicates that other effects, like image quality and module/cell types are important factors as well. 

We further find that AUROC is generally lower when using plant A or B as target as opposed to plants E or F. Possible explanations for this are the larger number of anomalies and the presence of sub-domains in plants A and B (see fig.~\ref{fig:dataset_umap_embedding}) which make the accurate prediction of anomalies harder. 

Note, that we do not report results for training on plant E as the contrastive loss did not converge. This is due to the lower fraction of anomalies in plant E, resulting in batches with only very few anomalous images. A larger batch size or special sampling strategy could have solved this issue.

\subsection{Hyperparameter Selection of the $k$\nobreakdash-NN Classifier}
\label{sec:hyperparameter_selection_for_the_knn_classifier}

The absence of labelled target images renders hyperparameter tuning of the $k$\nobreakdash-NN classifier on the target plant impossible. Thus, for practical applications, it is important that the $k$\nobreakdash-NN classifier is insensitive to the choice of hyperparameters.

\begin{figure}[tbp]
     \captionsetup[subfigure]{aboveskip=0pt, belowskip=0pt, justification=centering}
     \centering
     \begin{subfigure}[t]{\columnwidth}
         \centering
         \includegraphics[scale=0.5]{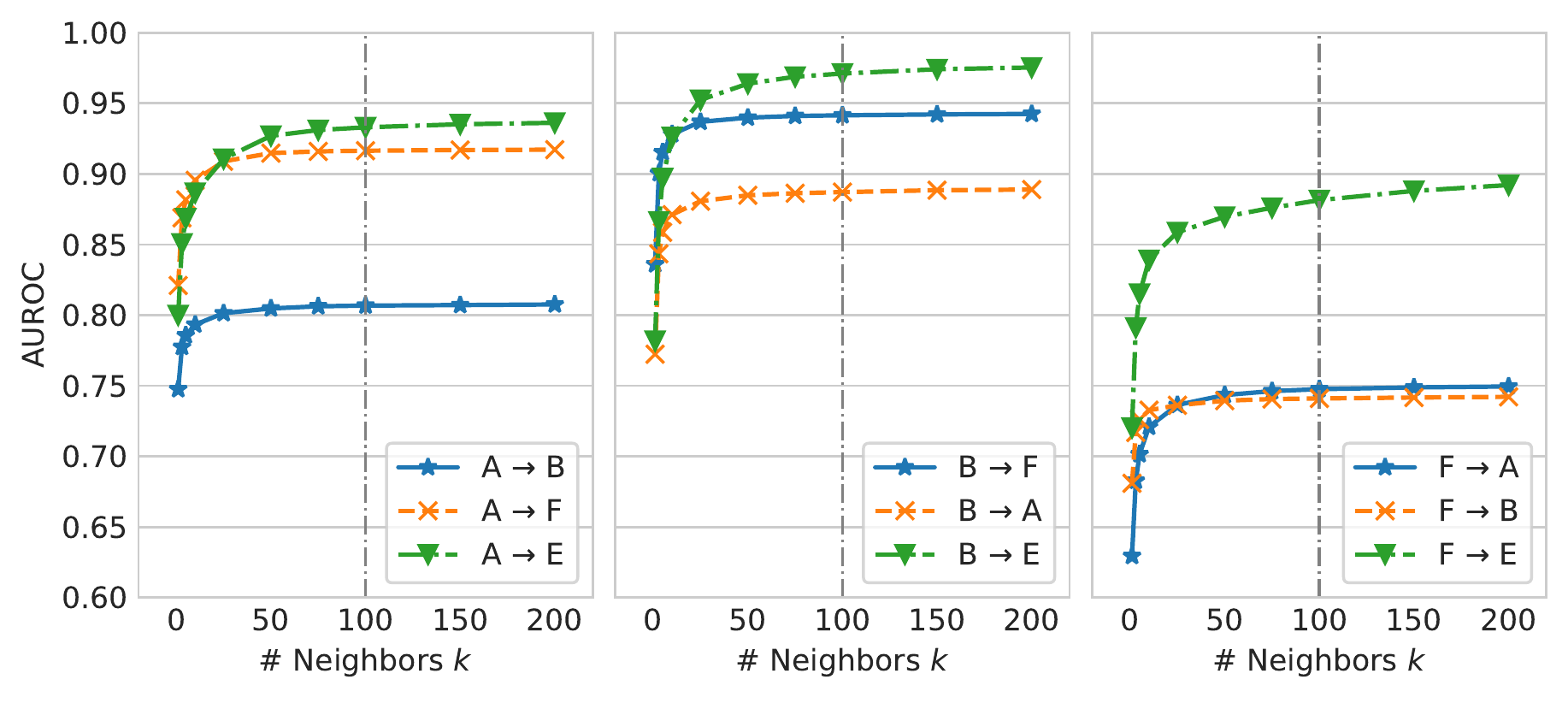}
         \caption{AUROC versus numbers $k$ of neighbors.}\label{fig:knn_n_neighbors_experiment}
     \end{subfigure}
     \par\smallskip
     \begin{subfigure}[t]{\columnwidth}
         \centering
         \includegraphics[scale=0.5]{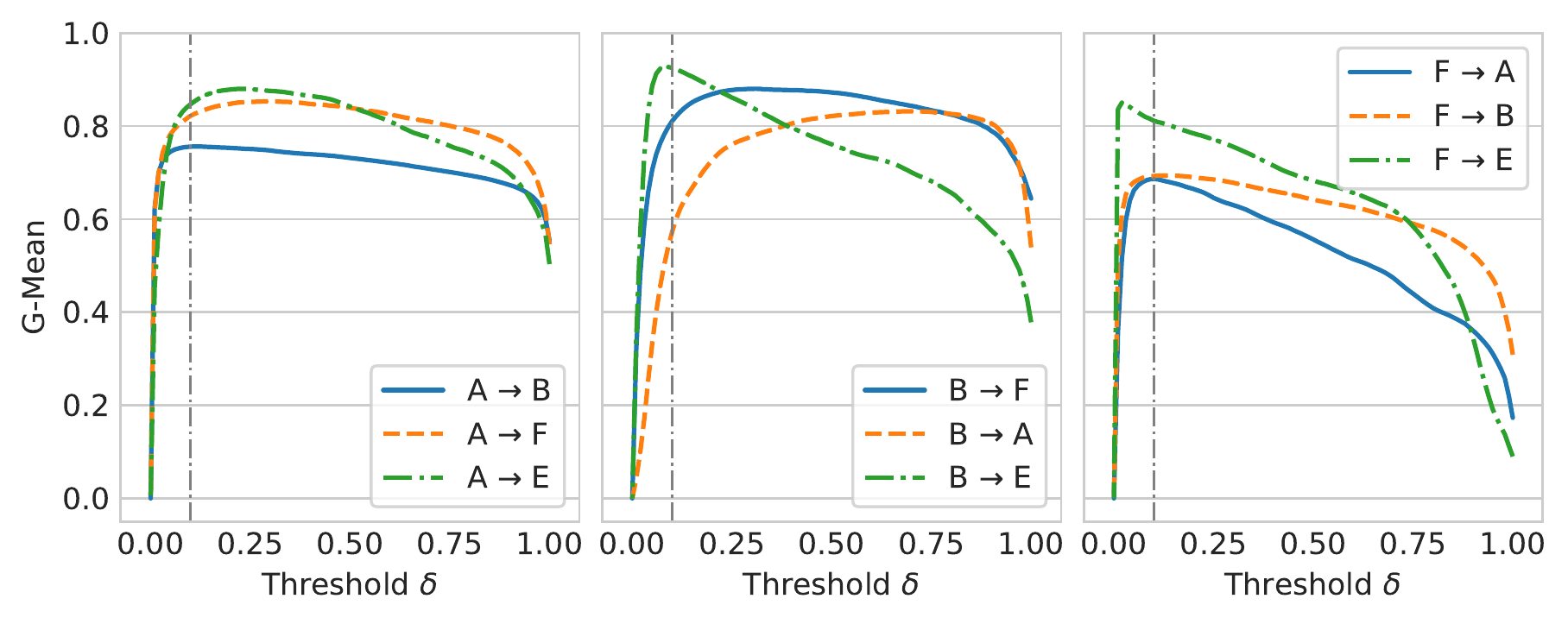}
         \caption{G\nobreakdash-Mean versus decision threshold $\delta$.}\label{fig:decision_threshold_selection}
     \end{subfigure}
        \caption{Prediction performance of the $k$\nobreakdash-NN classifier for different settings of the hyperparameters $k$ and $\delta$. The dashed vertical lines at $k = 100, \delta=0.1$ represent the trade-offs we use in practice. All classifiers are trained on contrastive embeddings of the model with random seed $1$ and best target AUROC on the respective task.}
        \label{fig:knn_classifier_hyperparameters}
\end{figure}

Fig.~\ref{fig:knn_n_neighbors_experiment} shows the $k$\nobreakdash-NN classifier AUROC for different numbers of neighbors. For all tasks the $k$\nobreakdash-NN classifier is insensitive to the choice of $k$ once it exceeds $25$. For some tasks the AUROC is still slightly increasing at $k=200$. However, as runtime also increases, we choose $k=100$ as trade-off in our experiments.

Another important hyperparameter is the decision threshold $\delta$, which is the fraction of anomalies required in the set of neighbors $\mathcal{N}_k$ to classify a target image as anomalous. Fig.~\ref{fig:decision_threshold_selection} shows the geometric mean (G\nobreakdash-Mean) of true positive rate and false positive rate for various decision thresholds $\delta$. While the classifier is more sensitive to the choice of $\delta$ (as compared to $k$), it behaves consistent across the tasks, taking on a high value for small thresholds. We choose $\delta = 0.1$ in practice to account for the imbalance between normal and anomalous images.

\subsection{Which Faults are Misclassified?}
\label{sec:which_anomalies_are_misclassified}

\begin{figure}[tbp]
\centering
\includegraphics[width=\columnwidth]{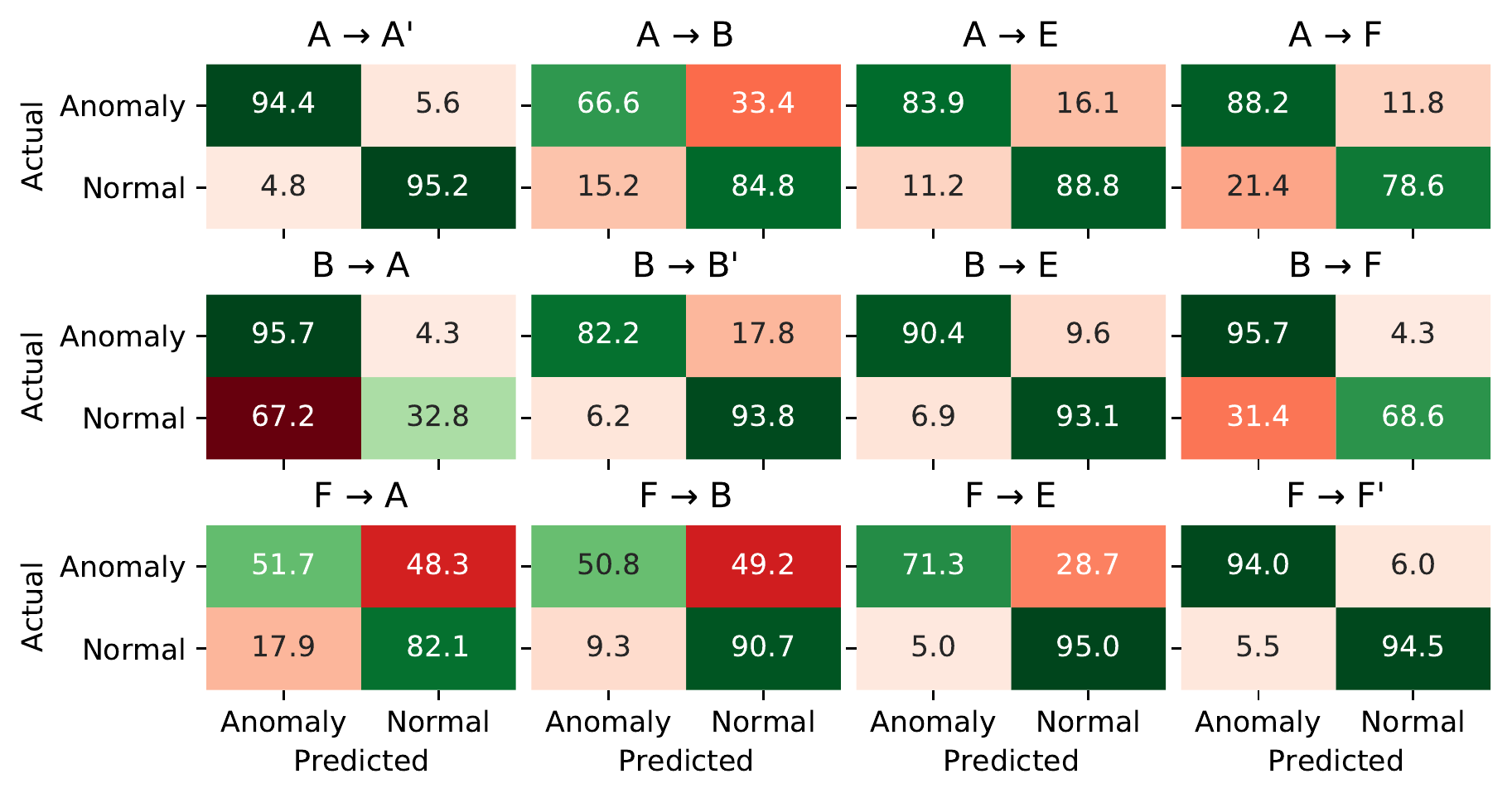}
\caption{Normalized confusion matrices of the $k$\nobreakdash-NN classifier ($k=100$ and $\delta=0.1$) on different tasks. We report averages using the model with best target AUROC at each random seed.}
\label{fig:confusion_matrices} 
\end{figure}

Using the hyperparameter settings from above, we make predictions with our contrastive $k$\nobreakdash-NN classifier and show the resulting confusion matrices in fig.~\ref{fig:confusion_matrices}. Averaged over all tasks, the fractions of correctly classified normal and anomalous images are \SI{79.4}{\percent} and \SI{77.1}{\percent}, respectively. Furthermore, the fraction of anomalies misclassified as normal is only \SI{22.9}{\percent} on average. Higher misclassification rates for the model trained on plant F suggest (in line with the results from sec.~\ref{sec:results_of_the_contrastive_knn_classifier}) that plant F is a poor choice for training. The less critical fraction of normal images misclassified as anomalous is \SI{20.6}{\percent} on average. An outlier in this metric is task B \textrightarrow{} A, which would require a higher decision threshold as can be seen in fig.~\ref{fig:decision_threshold_selection}.

For the purpose of analysis we have access to fine-grained target labels. Thus, we can analyze, which specific anomaly classes are misclassified, allowing us to identify potential systematic errors. Tab.~\ref{tab:which_defects_are_misclassified} reports our findings. With a few exceptions, error rates are below \SI{15}{\percent} for faults Mp, Sh, Sp, Pid, Cm+, Cs+ and C when training on plants A and B. 

For homogeneously overheated modules (Mh), we observe a high error rate. This is caused by the image-wise normalization applied during preprocessing and may be addressed in future works. High error rates also occur for D and Chs faults due to their small spatial extent in the image. This is a typical problem of convolutional neural networks. However, as D and Chs faults are not critical, we can accept the higher error rates. Interestingly, the model trained on plant F correctly identifies many Pid modules, despite the lack of Pid training examples in F. The model most likely transfers knowledge from the visually similar Mp class. This fails for the visually more unique Sh anomaly, of which plant F contains only $13$ examples.

\begin{table}[tbp]
\centering
\caption{Percentage of target anomalies misclassified as normal by our $k$\nobreakdash-NN classifier ($k=100$ and $\delta=0.1$) grouped by fault class. Fault criticality decreases from left to right. Error rates below \SI{15}{\percent} are green, between \SI{15}{\percent} and \SI{50}{\percent} orange and above \SI{50}{\percent} red. We report averages over three runs using the model with best target AUROC at each run.}
\label{tab:which_defects_are_misclassified}
\addtolength{\tabcolsep}{-1pt}
\definecolor{goodgreen}{rgb}{0.0, 0.5019607843137255, 0.0}
\newcommand{\good}[1]{\textcolor{goodgreen}{#1}}
\newcommand{\medium}[1]{\textcolor{orange}{#1}}
\newcommand{\bad}[1]{\textcolor{red}{#1}}
\begin{tabular}{l
S[table-format=2.1]
S[table-format=2.1]
S[table-format=2.1]
S[table-format=2.1]
S[table-format=2.1]
S[table-format=2.1]
S[table-format=2.1]
S[table-format=2.1]
S[table-format=2.1]
S[table-format=2.1]
S[table-format=2.1]}
\toprule
{Task} & \multicolumn{11}{c}{Actual Fault Class}\\ \cmidrule(lr){2-12}
 & {Mh} & {Mp} & {Sh} & {Sp} & {Pid} & {Cm+} & {Cs+} & {C} & {D} & {Chs} & {All}\\\midrule
A \textrightarrow{} A' & \good{7.8} & \good{0.0} & \good{0.0} & \good{11.2} & \good{0.3} & {--} & \good{0.0} & \good{3.4} & \medium{16.6} & \medium{28.4} & \good{5.6}\\
A \textrightarrow{} B & \bad{76.6} & \good{0.3} & \good{1.3} & \good{0.0} & \good{1.5} & \good{3.6} & \good{2.7} & \good{13.5} & \bad{55.8} & \medium{33.0} & \medium{33.4}\\
A \textrightarrow{} E & {--} & \good{0.0} & \good{2.7} & \good{0.0} & {--} & \good{0.0} & \good{0.1} & \medium{18.6} & {--} & \bad{72.0} & \medium{16.1}\\
A \textrightarrow{} F & \good{0.0} & \medium{16.1} & \good{0.0} & \good{2.1} & {--} & \good{1.2} & \good{4.9} & \good{11.9} & \bad{55.3} & {--} & \good{11.8}\\\midrule
B \textrightarrow{} A & \good{0.0} & \good{0.0} & \good{0.1} & \good{0.0} & \good{0.7} & \good{0.0} & \good{0.0} & \good{3.8} & \good{10.8} & \good{9.4} & \good{4.3}\\
B \textrightarrow{} B' & \medium{37.0} & \good{0.0} & \good{3.9} & \good{0.0} & \good{1.1} & \medium{35.0} & \good{0.3} & \good{11.2} & \bad{97.4} & \medium{21.7} & \medium{17.8}\\
B \textrightarrow{} E & {--} & \good{0.0} & \good{8.7} & \good{0.1} & {--} & \good{0.0} & \good{3.9} & \good{10.3} & {--} & \bad{51.9} & \good{9.6}\\
B \textrightarrow{} F & \good{0.0} & \good{0.0} & \good{0.0} & \good{0.0} & {--} & \good{0.0} & \good{0.2} & \good{3.9} & \bad{66.0} & {--} & \good{4.3}\\\midrule
F \textrightarrow{} A & \medium{30.4} & \good{0.0} & \bad{60.5} & \good{0.0} & \medium{27.0} & \good{0.0} & \good{1.3} & \bad{62.1} & \bad{62.1} & \bad{59.8} & \medium{48.3}\\
F \textrightarrow{} B & \bad{91.2} & \good{0.0} & \bad{71.8} & \good{1.0} & \good{8.6} & \good{3.6} & \good{1.0} & \medium{26.6} & \medium{30.8} & \bad{75.8} & \medium{49.2}\\
F \textrightarrow{} E & {--} & \good{7.0} & \bad{68.9} & \good{2.3} & {--} & \medium{22.3} & \good{2.3} & \medium{32.6} & {--} & \bad{84.2} & \medium{28.7}\\
F \textrightarrow{} F' & {--} & {--} & {--} & \good{0.0} & {--} & \good{0.7} & \good{0.3} & \good{6.7} & {--} & \good{0.0} & \good{6.0}\\\bottomrule
\end{tabular}
\end{table}

\subsection{Visualization of Misclassified IR Images}

\begin{figure}[tbp]
\centering
\includegraphics[width=\columnwidth]{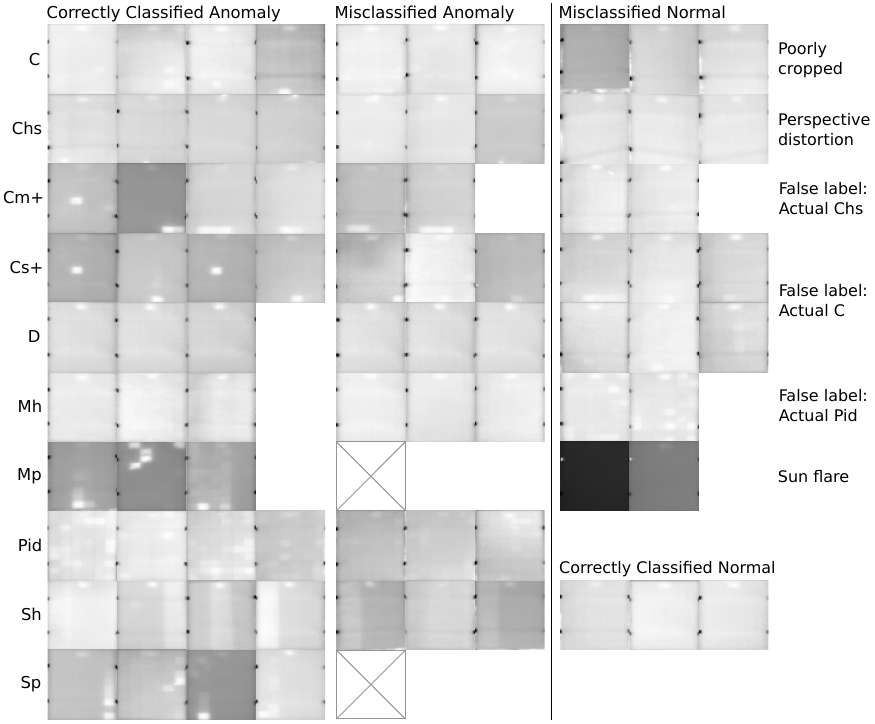}
\caption{Exemplary predictions of our $k$\nobreakdash-NN classifier ($k=100$ and $\delta=0.1$) for IR images of plant B by the model trained on plant A. We use the model at seed $1$ with best target AUROC and show preprocessed patches.}
\label{fig:prediction_collage} 
\end{figure}

To build an intuition for the quantitative results of our method, we make predictions on IR images and visualize both correct and false predictions in fig.~\ref{fig:prediction_collage}.

As shown by the examples, the high misclassification rates for the Mh, D and Chs anomalies can be explained by their high visual similarity to the normal images. Similarly, we find that primarily those anomalous images are misclassified that exhibit lower local temperature differences and are visually more similar to the normal images. This is a good indicator for the smoothness of the learned contrastive representations and thus the robustness of our approach.

Fig.~\ref{fig:prediction_collage} also hightlights a few misclassified normal images. Interestingly, most of these are valid anomalies with false ground truth labels. There are also cases of poorly cropped images, images with strong perspective distortion or images with sun reflections. Our method correctly identifies them as anomalies despite never having been trained on such examples.

\subsection{Embedding Visualizations}

\begin{figure}[!t]
     \captionsetup[subfigure]{aboveskip=2pt, belowskip=0pt, labelformat=empty, justification=centering}
     \newcommand\sizefactor{0.21}
     \centering
     \begin{subfigure}[t]{\sizefactor\columnwidth}
         \centering
         \includegraphics[width=\textwidth,angle=270,origin=c]{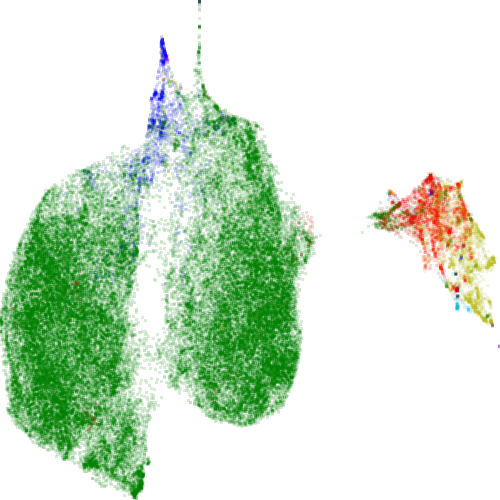}
        \caption{A \textrightarrow{} A'}
     \end{subfigure}\hfill
     \begin{subfigure}[t]{\sizefactor\columnwidth}
         \centering
         \includegraphics[width=\textwidth]{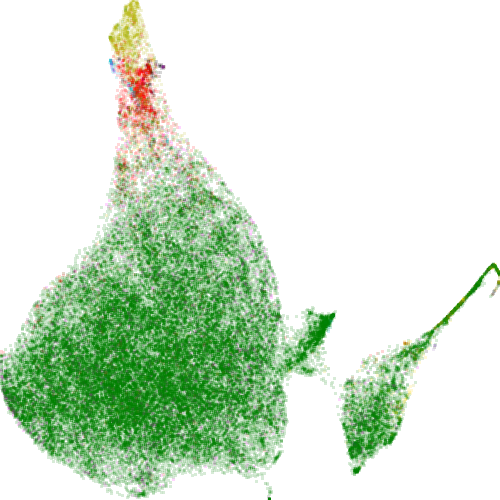}
         \caption{A \textrightarrow{} B}
     \end{subfigure}\hfill
     \begin{subfigure}[t]{\sizefactor\columnwidth}
         \centering
         \includegraphics[width=\textwidth]{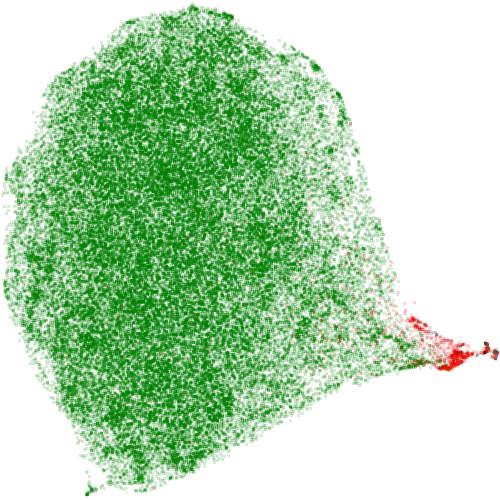}
         \caption{A \textrightarrow{} E}
     \end{subfigure}\hfill
     \begin{subfigure}[t]{\sizefactor\columnwidth}
         \centering
         \includegraphics[width=\textwidth]{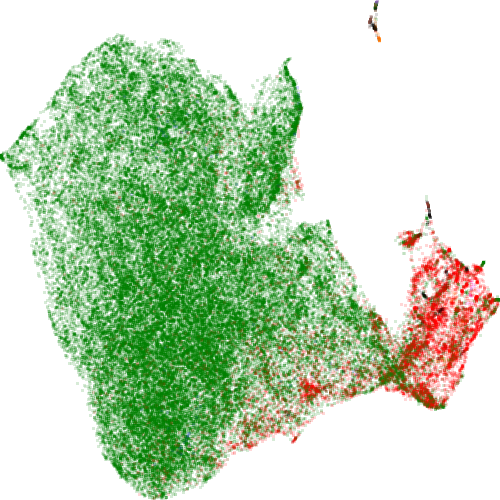}
         \caption{A \textrightarrow{} F}
     \end{subfigure}
     \par\smallskip
     \begin{subfigure}[t]{\sizefactor\columnwidth}
         \centering
         \includegraphics[width=\textwidth]{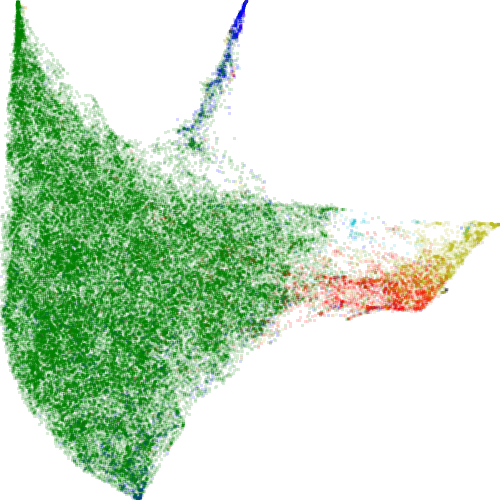}
         \caption{B \textrightarrow{} A}
     \end{subfigure}\hfill
     \begin{subfigure}[t]{\sizefactor\columnwidth}
         \centering
         \includegraphics[width=\textwidth]{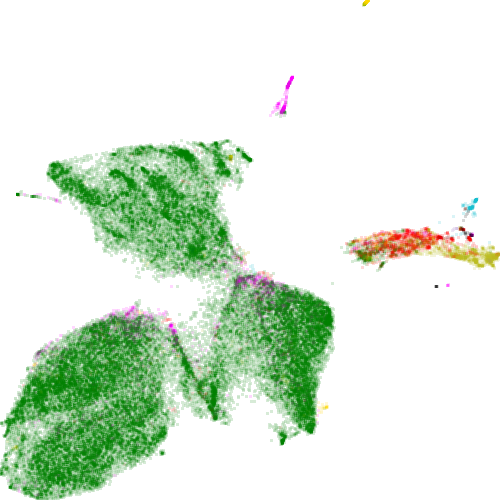}
         \caption{B \textrightarrow{} B'}
     \end{subfigure}\hfill
     \begin{subfigure}[t]{\sizefactor\columnwidth}
         \centering
         \includegraphics[width=\textwidth]{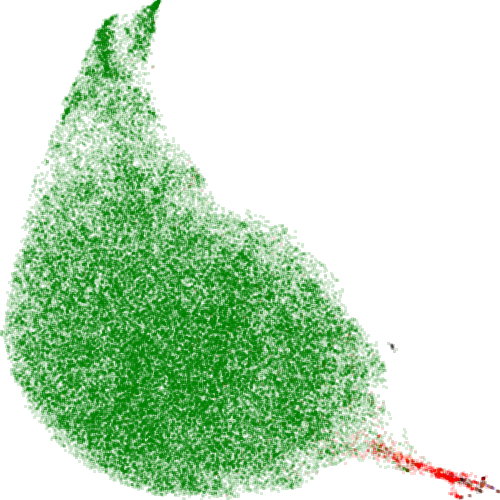}
         \caption{B \textrightarrow{} E}
     \end{subfigure}\hfill
     \begin{subfigure}[t]{\sizefactor\columnwidth}
         \centering
         \includegraphics[width=\textwidth]{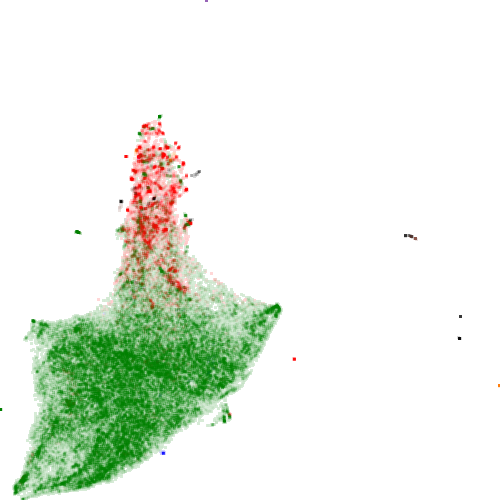}
         \caption{B \textrightarrow{} F}
     \end{subfigure}
     \par\smallskip
     \begin{subfigure}[t]{\sizefactor\columnwidth}
         \centering
         \includegraphics[width=\textwidth]{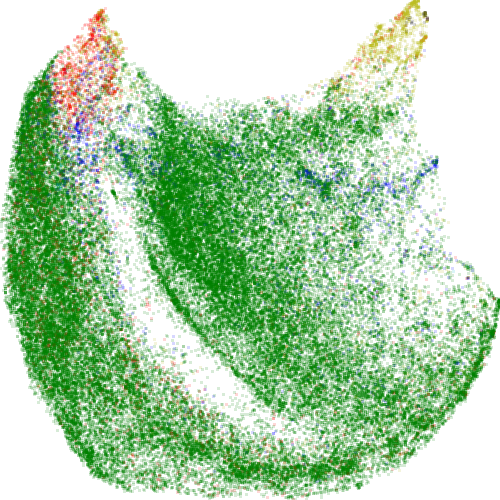}
         \caption{F \textrightarrow{} A}
     \end{subfigure}\hfill
     \begin{subfigure}[t]{\sizefactor\columnwidth}
         \centering
         \includegraphics[width=\textwidth]{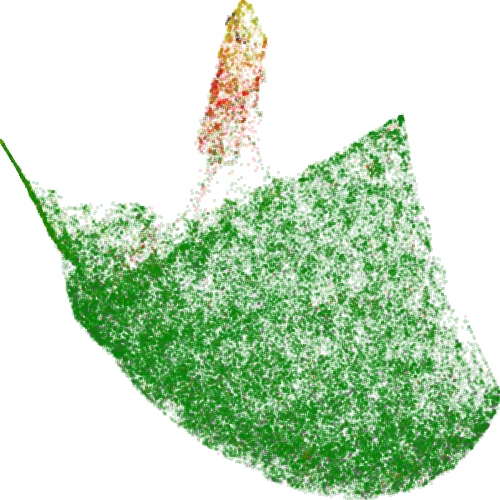}
         \caption{F \textrightarrow{} B}
     \end{subfigure}\hfill
     \begin{subfigure}[t]{\sizefactor\columnwidth}
         \centering
         \includegraphics[width=\textwidth]{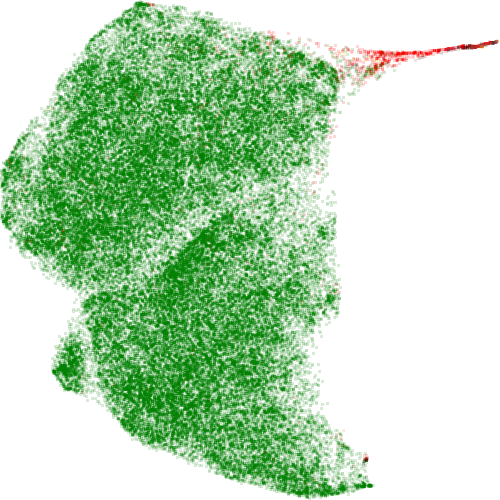}
         \caption{F \textrightarrow{} E}
     \end{subfigure}\hfill
     \begin{subfigure}[t]{\sizefactor\columnwidth}
         \centering
         \includegraphics[width=\textwidth]{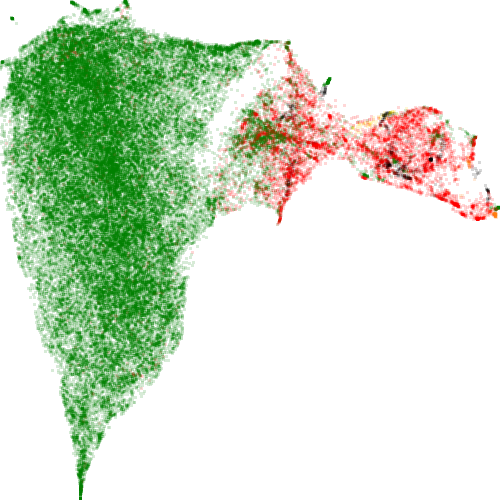}
         \caption{F \textrightarrow{} F'}
     \end{subfigure}
     \par\smallskip
     \begin{subfigure}[t]{\columnwidth}
         \centering
         \includegraphics[width=\textwidth]{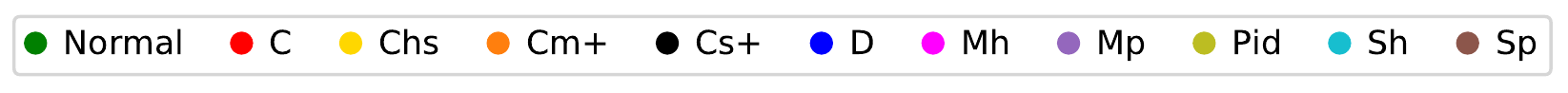}
     \end{subfigure}
        \caption{UMAP projections (with $50$ neighbors per sample and minimum distance of $0.1$) of the target datasets embedded by ResNet\nobreakdash-$34$ after supervised contrastive training. Embeddings are obtained behind the ResNet\nobreakdash-$34$ average pooling layer. For each task the model at seed $1$ with best target AUROC is shown.}
        \label{fig:result_embeddings}
\end{figure}

As another means to interpret our models, we visualize the representations learned by supervised contrastive training in fig.~\ref{fig:result_embeddings}. Here, for most tasks, the representations clearly separate normal and anomalous images, which explains the overall high AUROC and AP scores achieved. Exceptions are tasks F \textrightarrow{} A and F \textrightarrow{} B, where many anomalies lie within the normal cluster resulting in a low recall. We can also see that the anomaly classes Mp, Sh, Sp, Pid, Cm+ and Cs+, which achieved low error rates in sec.~\ref{sec:which_anomalies_are_misclassified}, have a larger distance to the normal modules than anomalies with higher error rates (Mh, D, Chs). The C anomalies often lie somewhere in between, which is in accordance to the slightly higher error rates of around \SI{10}{\percent}.

\subsection{Detection of Unknown Anomalies}
\label{sec:detection_of_unknown_anomalies}

One goal of our method is the ability to reliably detect anomalies in the target dataset, which are not contained in the source dataset. To analyze how well our method deals with such unknown anomalies, we remove all anomalies of classes Mp, Sh, Sp, Cm+ and Cs+ from the source datasets of plants A and B, retrain our models and evaluate on the full target datasets containing all anomaly classes. We chose precisely these classes, as they make up only \SI{3.1}{\percent} and \SI{5.2}{\percent} of all anomalies in datasets A and B. This leaves dataset sizes nearly unchanged, providing us with a more comparable result. For most tasks, the resulting target AUROCs (see tab.~\ref{tab:leaveout_classes_results}) do not deviate much from the respective AUROCs of the models trained on all anomaly classes. Similarly, we do not observe any change in model convergence during training, as shown in fig.~\ref{fig:ap_over_training_leaveout_classes} in appendix \ref{sec:additional_comparisons_for_training_without_some_anomalies}. The results indicate that our method can reliably detect unknown anomalies.

\begin{table}[tbp]
\centering
\caption{Target AUROCs of our contrastive $k$\nobreakdash-NN classifier trained on datasets where anomaly classes Mp, Sh, Sp, Cm+ and Cs+ are left out versus the baseline trained on the full dataset. All values are averages over three training runs.}
\label{tab:leaveout_classes_results}
\begin{tabular}{l
S[table-format=2.2]
S[table-format=2.2]
l
S[table-format=2.2]
S[table-format=2.2]}
\toprule
Task & \multicolumn{2}{c}{Variant} & Task & \multicolumn{2}{c}{Variant}\\\cmidrule(lr){2-3}\cmidrule(lr){5-6}
 & {Full dataset} & {Leaveout} & & {Full dataset} & {Leaveout}\\\midrule
A \textrightarrow{} A' & 98.39 & 98.29 & B \textrightarrow{} A  & 86.86 & 83.43\\
A \textrightarrow{} B  & 80.38 & 80.60 & B \textrightarrow{} B' & 93.38 & 93.38\\
A \textrightarrow{} E  & 91.93 & 92.26 & B \textrightarrow{} E  & 96.64 & 96.67\\
A \textrightarrow{} F  & 91.06 & 91.63 & B \textrightarrow{} F  & 94.35 & 93.92\\\bottomrule
\end{tabular}
\end{table}

\subsection{Comparison with Cross-Entropy Classifier}
\label{sec:comparison_with_cross_entropy_classifier}

We compare our method with a deep convolutional binary classifier based on ResNet\nobreakdash-$34$, which is trained with standard cross-entropy loss using the same data preprocessing, data augmentation and training settings as our method (see sec.~\ref{sec:implementation_details}). While the convolutional backbone is identical to our contrastive model, a softmax-activated fully connected layer with $2$ outputs is used on top of the $2$D global average pooling layer. A projection head is not employed. 

As shown in tab.~\ref{tab:results_contrastive_vs_supervised}, our method outperforms the cross-entropy classifier in terms of target AUROC in many cases which is in accordance to the literature \cite{Khosla.2020}. Only on tasks F \textrightarrow{} B and F \textrightarrow{} E our method falls behind. This could be due to the smaller dataset size and thus smaller absolute number of anomalies in plant F. It indicates that our method is more sensitive to the dataset size, i.e., is less accurate on small datasets but profits more from larger dataset sizes than the cross-entropy classifier.

The same result is reflected in the AP, which is exemplary shown for plant A over the course of the training in fig.~\ref{fig:contrastive_vs_cross_entropy_ABFE}. An additional analysis for plants B and F is provided in Appendix \ref{sec:additional_comparisons_with_the_cross_entropy_classifier}. Furthermore, we find that due to the large size of our datasets target AP converges within a single training epoch.

\begin{figure*}[tbp] 
\centering
\includegraphics[width=\textwidth]{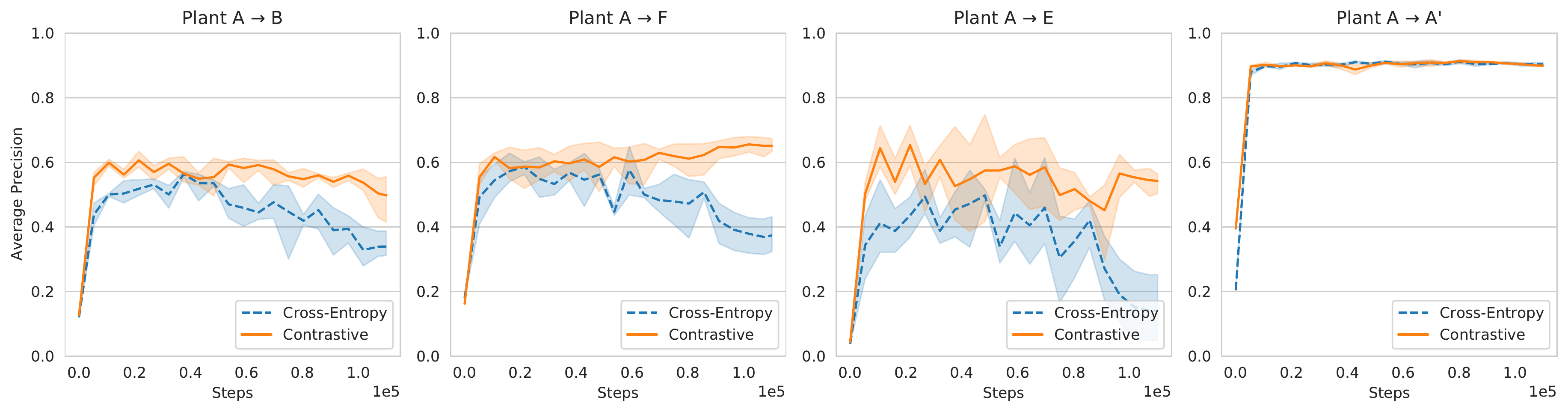}
\caption{Average precision over the course of training of our contrastive $k$\nobreakdash-NN classifier (orange line) versus a supervised binary classifier trained on cross-entropy loss (dashed blue line). Plant A is used as source. Shaded regions indicate the \SI{95}{\percent} confidence interval over three runs.}
\label{fig:contrastive_vs_cross_entropy_ABFE} 
\end{figure*}

\begin{table}[tbp]
\centering
\caption{Target AUROCs of our contrastive $k$\nobreakdash-NN classifier versus a binary classifier trained with cross-entropy loss. We report the best values achieved and values for models selected via two different validation datasets (Val $0$ and Val $1$). All values are averages over three training runs. Values of the better method are in bold.}
\label{tab:results_contrastive_vs_supervised}
\addtolength{\tabcolsep}{-0.5pt}
\begin{tabular}{lcc
S[table-format=2.2]
S[table-format=2.2]
S[table-format=2.2]
S[table-format=2.2]
S[table-format=2.2]
S[table-format=2.2]}
\toprule
{Task} & {Val $0$} & {Val $1$} & \multicolumn{3}{c}{Contrastive AUROC} & \multicolumn{3}{c}{Cross-Entropy AUROC}\\ \cmidrule(lr){4-6}\cmidrule(lr){7-9}
 & & & {@Val $0$} & {@Val $1$} & {Best} & {@Val $0$} & {@Val $1$} & {Best}\\\midrule
A \textrightarrow{} A' & -- & -- & {--} & {--} & 98.39 & {--} & {--} & \bfseries{98.64}\\
A \textrightarrow{} B & F & E & \bfseries{78.83} & \bfseries{79.62} & \bfseries{80.38} & 74.68 & 77.76 & 78.19\\
A \textrightarrow{} E & F & B & \bfseries{90.85} & \bfseries{90.02} & \bfseries{91.93} & 81.45 & 86.49 & 88.71\\
A \textrightarrow{} F & B & E & \bfseries{88.82} & \bfseries{90.49} & \bfseries{91.06} & 85.18 & 86.10 & 89.26\\\midrule
B \textrightarrow{} A & F & E & \bfseries{77.01} & \bfseries{80.25} & \bfseries{86.86} & 68.35 & 69.95 & 76.81\\
B \textrightarrow{} B' & -- & -- & {--} & {--} & 93.38 & {--} & {--} & \bfseries{95.45}\\
B \textrightarrow{} E & F & A & \bfseries{95.20} & \bfseries{93.20} & 96.64 & 92.54 & 88.32 & \bfseries{96.66}\\
B \textrightarrow{} F & A & E & 88.13 & 92.42 & \bfseries{94.35} & \bfseries{89.94} & \bfseries{92.51} & 93.85\\\midrule
F \textrightarrow{} A & B & E & \bfseries{66.14} & \bfseries{61.58} & 73.29 & 63.46 & 55.49 & \bfseries{75.22}\\
F \textrightarrow{} B & A & E & 69.22 & 72.68 & 74.54 & \bfseries{74.91} & \bfseries{76.83} & \bfseries{77.52}\\
F \textrightarrow{} E & A & B & 84.06 & 86.42 & 88.86 & \bfseries{88.02} & \bfseries{88.65} & \bfseries{90.64}\\
F \textrightarrow{} F' & -- & -- & {--} & {--} & 97.44 & {--} & {--} & \bfseries{97.54}\\\bottomrule
\end{tabular}
\end{table}

\subsection{Module-level Aggregation of Predictions}
\label{sec:module_level_aggregation_of_predictions}

\begin{figure}[tbp]
\centering
\includegraphics[width=\columnwidth]{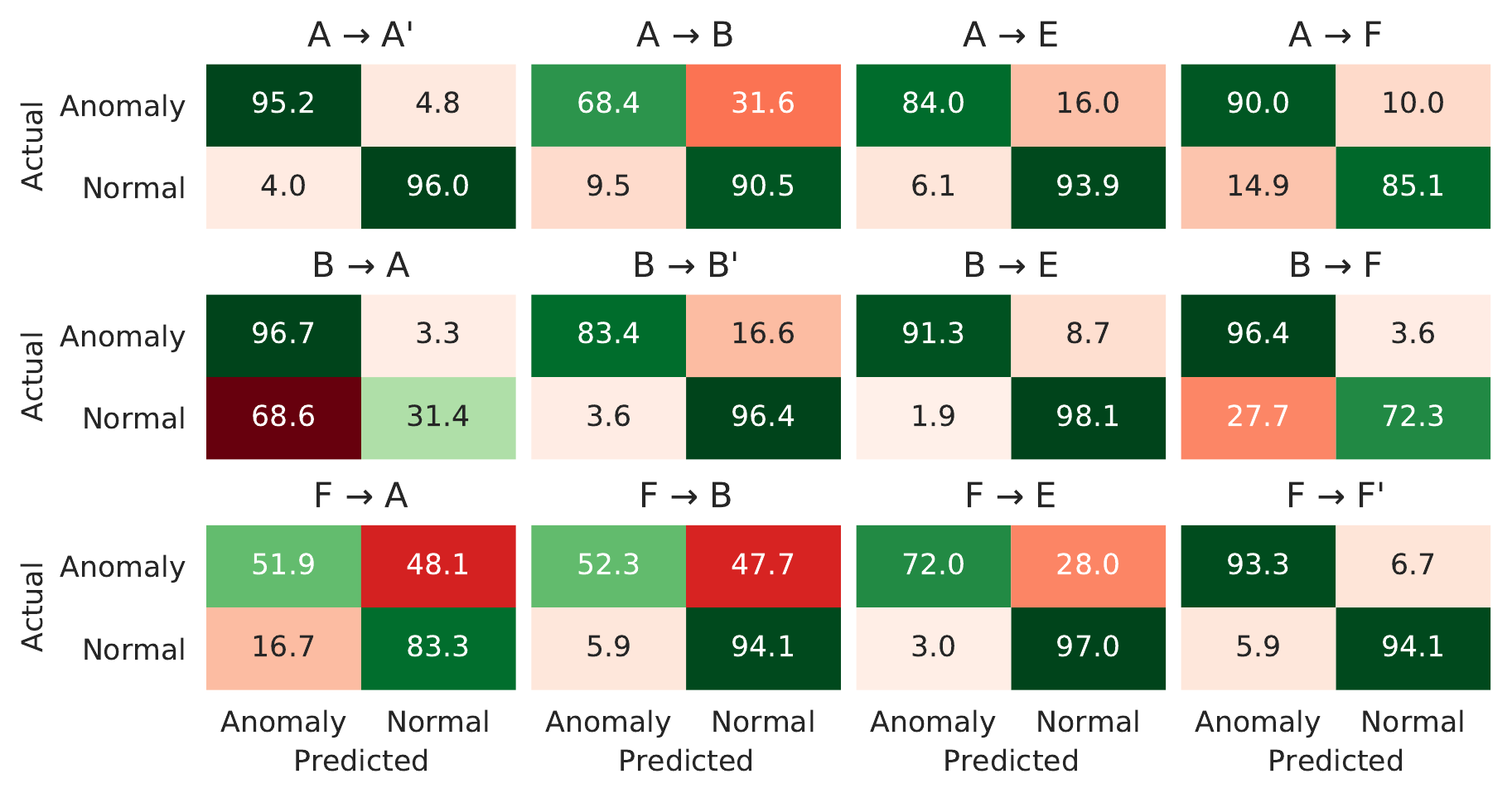}
\caption{Normalized confusion matrices of the $k$\nobreakdash-NN classifier ($k=100$ and $\delta=0.1$) for predictions aggregated on module-level. We report averages using the model with best target AUROC at each random seed.}
\label{fig:confusion_matrices_module_level} 
\end{figure}

As there are on average $39.4$ IR images of each PV module we can aggregate predictions of those images to obtain a final prediction for the module. Specifically, we predict a module as anomalous if at least one half of the corresponding images are predicted anomalous. As indicated by the resulting confusion matrices in fig.~\ref{fig:confusion_matrices_module_level}, on average \SI{82.9}{\percent} of all normal and \SI{78.1}{\percent} of all anomalous modules are correctly classified. On average, \SI{17.1}{\percent} of the normal modules are misclassified as anomalous and \SI{21.9}{\percent} of the anomalous modules are misclassified as normal. As compared to the image-level predictions (see sec.~\ref{sec:which_anomalies_are_misclassified}) module-level aggregation improves especially upon the detection rate of normal modules, but also yields a one percent higher detection rate for anomalies. These results suggest that the hierarchical structure of our dataset is beneficial for the accurate detection of anomalous PV modules.

\subsection{Exemplary Application to Labelling of IR Datasets}

For the development of future fault classification methods large IR image datasets are needed. Our method can drastically reduce the time and effort needed for labelling such datasets by automatically rejecting the majority of normal (and thus uninteresting) PV modules. For example, when labelling plant E, one would have to manually sight $14662$ PV modules, of which only $296$ are anomalous, i.e., actually interesting. Applying our method (trained for example on plant B) could automatically reject \SI{98.1}{\percent} of the normal modules, leaving only $273$ normal modules for manual sighting. The cost for this improvement is the loss of $26$ anomalous modules, which are misclassified as normal. In total, $543$ modules are left for manual sighting. Assuming an expert takes three seconds to label one module, this reduces the time needed to label plant E from $12.2$~hours to only $27$ minutes. Adjusting the decision threshold during module-level aggregation allows to trade off lost anomalies and time savings.


\section{Discussion and Conclusion}
\label{sec:conclusion}

\subsubsection{Summary}

In this work, we proposed a novel method for the detection of PV module faults in IR images using supervised contrastive learning. Instead of sampling train and test data from the same PV plant, we performed training with labelled IR images of one source plant and made predictions on another target plant. We identified domain shift between source and target data as a problem in this setting and addressed it by learning transferable representations with a supervised contrastive loss. A $k$\nobreakdash-NN classifier was used on top of these representations to detect unknown anomalies in the target plant. Experiments on nine different combinations of four source and target datasets showed the effectiveness of our method, which achieved an AUROC of \SI{73.3}{\percent} to \SI{96.6}{\percent} and even outperformed a binary cross-entropy classifier in some cases. We further found that our method converges quickly and is relatively insensitive to hyperparameter settings, making it well suited for practical applications. Using fault labels for $10$ different types of anomalies, we found that our method most frequently misses anomalies with a small spatial extent in the image, e.g. overheated bypass diodes or small hot spots. Most striking, our method showed no significant drop in AUROC after removing five of the ten anomaly classes from the training datasets, proving its ability to reliably detect unknown anomalies. Finally, we improved detection accuracy by aggregating predictions of multiple IR images belonging to the same PV module.

\subsubsection{Practical Relevance}

Increasing PV deployments and aging PV plants require regular inspections to ensure a safe operation and maximum power output, yield and profitability of a plant. The large size of most PV plants and potentially high labour cost renders a manual inspection economically infeasible and raises the need for fully automatic plant inspection. Our method is highly relevant for such inspection systems, as it automatically identifies anomalous PV modules in a large number of IR images. This enables targeted repairs and restoration of the original performance of a PV plant. Apart from the inspection of existing plants, automatic inspection is further useful for the commissioning of new plants.

One problem of existing fault detection methods is that they do not explicitly consider domain shift between different PV plants. This means a fault detector must be fine-tuned on labelled training images of each new PV plant that is inspected. This is not only labor-intensive, but also time-consuming, as training a neural network takes several hours. Opposed to that, our method explicitly handles domain shift. This way, it needs to be trained only once on a labelled dataset and generalizes afterwards to new PV plants without further fine-tuning. This is of major importance for realizing economically viable plant inspection systems that work for many different PV plants without the need for a time-consuming and costly setup phase.

Apart from automatic plant inspection, our method can also aid the manual labelling of IR datasets. This facilitates creation of large-scale datasets, which are needed for the development of the next generation of automatic fault detection algorithms.

\subsubsection{Future Works}

We presented a PV module fault detection method, which overcomes domain shift between different PV plants and generalizes beyond the training dataset without the need for huge amounts of labelled training data. While this is an important milestone, further measures could improve domain adaption and increase detection accuracy on new PV plants. For example, future works could explore active domain adaptation techniques, such as Maximum Mean Discrepancy. In addition, multi-domain adaptation, which uses multiple labelled source datasets from different PV plants simultaneously, could be taken into consideration.


\section{Acknowledgements}

We would like to thank Sanjay Venugopal for valuable discussions about the contrastive learning objective. We gratefully acknowledge the German Federal Ministry for Economic Affairs and Energy (BMWi) and the IBC SOLAR AG for financial funding of the project iPV4.0 (FKZ: 0324286). This work was financially supported by the State of Bavaria via the project PV-Tera (No. 446521a/20/5) and by BMWi via the project COSIMA (FKZ: 032429A). We sincerely thank the Allianz Risk Consulting GmbH / Allianz Zentrum für Technik (AZT) in Munich, Germany for supporting the project.


\FloatBarrier
\newpage



\newpage
\onecolumn

\appendix

\section{Appendix}

\subsection{Additional Comparisons with the Cross-Entropy Classifier}
\label{sec:additional_comparisons_with_the_cross_entropy_classifier}

\begin{figure*}[htbp]
     \captionsetup[subfigure]{aboveskip=0pt, belowskip=0pt, justification=centering}
     \centering
     \begin{subfigure}[t]{\textwidth}
         \centering
         \includegraphics[scale=0.5]{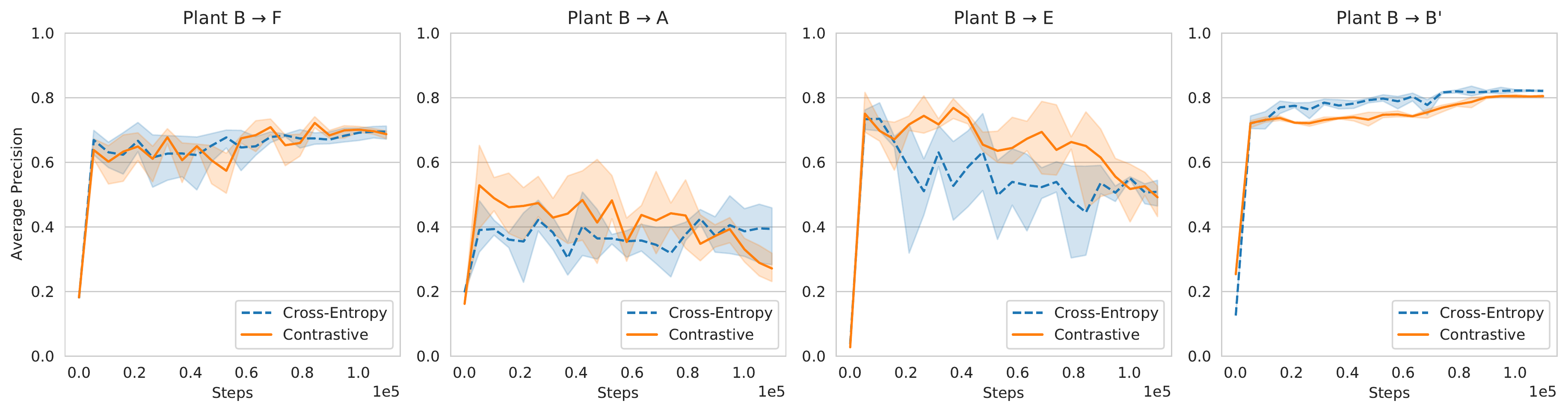}
     \end{subfigure}
     \par\smallskip
     \begin{subfigure}[t]{\textwidth}
         \centering
         \includegraphics[scale=0.5]{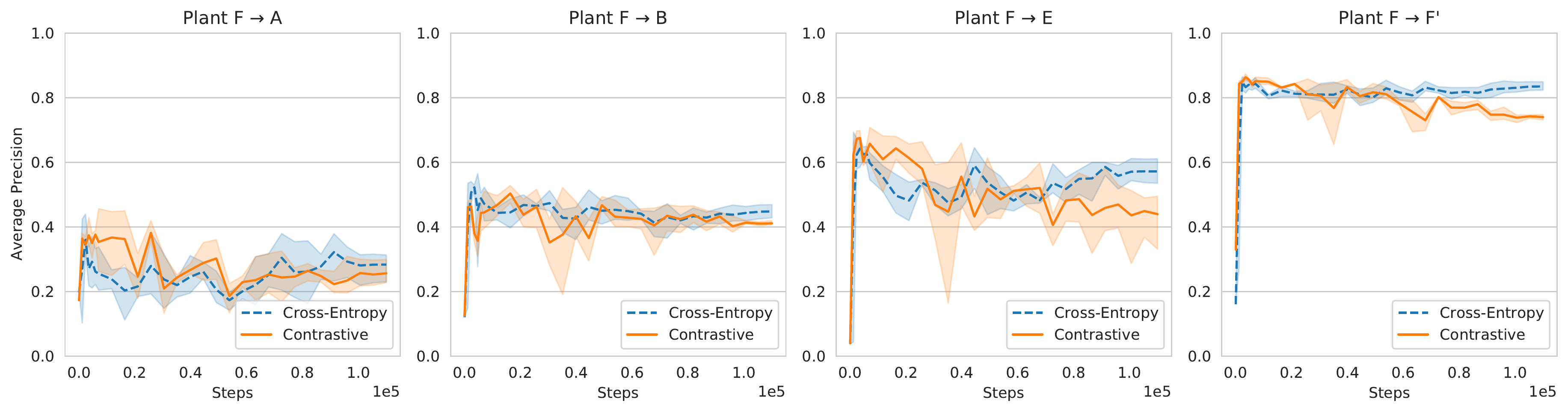}
     \end{subfigure}
        \caption{Average precision over the course of training of our contrastive $k$-NN classifier (orange line) versus a supervised binary classifier trained on cross-entropy loss (dashed blue line). The top row uses plant B as source, the bottom row plant F. Shaded regions indicate the \SI{95}{\percent} confidence interval over three runs.}
        \label{fig:contrastive_vs_cross_entropy_BFAE_FABE}
\end{figure*}

Fig.~\ref{fig:contrastive_vs_cross_entropy_BFAE_FABE} shows additional results for the comparison of our method with a cross-entropy classifier performed in sec.~\ref{sec:comparison_with_cross_entropy_classifier}.

\newpage

\subsection{Additional Comparisons for Training without some Anomalies}
\label{sec:additional_comparisons_for_training_without_some_anomalies}

\begin{figure*}[htbp]
     \captionsetup[subfigure]{aboveskip=0pt, belowskip=0pt, justification=centering}
     \centering
     \begin{subfigure}[t]{\textwidth}
         \centering
         \includegraphics[scale=0.5]{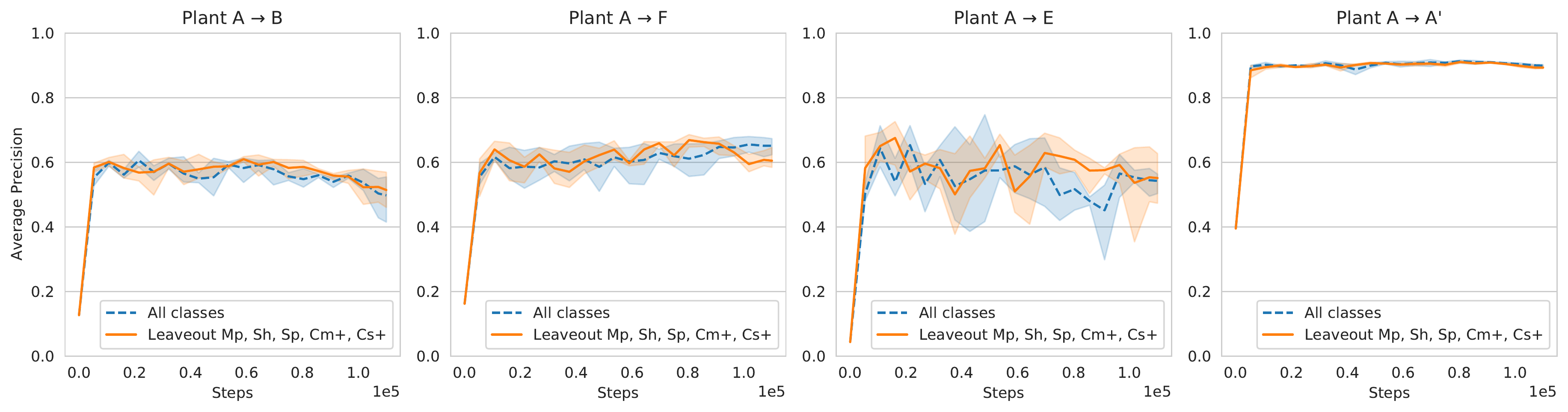}
     \end{subfigure}
     \par\smallskip
     \begin{subfigure}[t]{\textwidth}
         \centering
         \includegraphics[scale=0.5]{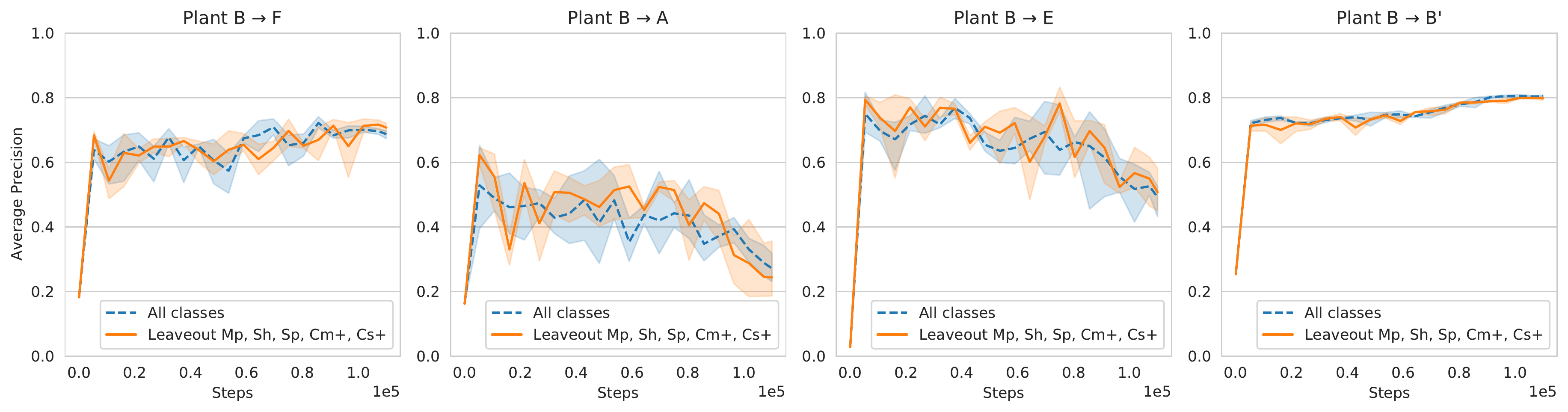}
     \end{subfigure}
        \caption{Average precision over the course of training of our contrastive $k$-NN classifier trained on datasets without anomaly classes Mp, Sh, Sp, Cm+ and Cs+ (orange line) versus the baseline trained on the full dataset (dashed blue line). Shaded regions indicate the \SI{95}{\percent} confidence interval over three runs.}
        \label{fig:ap_over_training_leaveout_classes}
\end{figure*}

Fig.~\ref{fig:ap_over_training_leaveout_classes} shows additional results for our method trained on reduced source datasets without Mp, Sh, Sp, Cm+ and Cs+ anomalies. It extends the results presented in sec.~\ref{sec:detection_of_unknown_anomalies}.

\end{document}